\documentclass{article}

\usepackage{iclr2027_conference,times}
\iclrfinalcopy

\usepackage{amsmath,amsfonts,bm}

\def\eqref#1{equation~\ref{#1}}

\def\1{\bm{1}}

\DeclareMathAlphabet{\mathsfit}{\encodingdefault}{\sfdefault}{m}{sl}
\SetMathAlphabet{\mathsfit}{bold}{\encodingdefault}{\sfdefault}{bx}{n}

\usepackage{booktabs}
\usepackage{graphicx}
\usepackage{wrapfig}
\usepackage{needspace}
\usepackage{placeins}
\usepackage{multirow}
\usepackage{amsmath,amssymb}
\usepackage{hyperref}
\usepackage{url}
\usepackage{subcaption}
\usepackage[most]{tcolorbox}

\definecolor{ariadneframe}{HTML}{8EABB5}
\definecolor{ariadneback}{HTML}{F4F7F8}
\definecolor{ariadnetitle}{HTML}{E7EFF1}
\newtcblisting{ariadneoutput}[1]{
  listing only,
  title={#1},
  colback=ariadneback,
  colframe=ariadneframe,
  colbacktitle=ariadnetitle,
  coltitle=black,
  fonttitle=\bfseries\small,
  boxrule=0.5pt,
  arc=1.5pt,
  left=1mm,
  right=1mm,
  top=1mm,
  bottom=1mm,
  listing options={
    basicstyle=\ttfamily\footnotesize,
    columns=fullflexible,
    keepspaces=true,
    breaklines=true,
    showstringspaces=false
  }
}

\title{Knossos and Ariadne: Benchmarking and Learning Complete Diagram Topology Extraction with Vision–Language Models}

\author{\begin{minipage}{\dimexpr\textwidth-2\tabcolsep\relax}
\vspace{1.2em}
\centering
Bangwei Guo\textsuperscript{1}, Xujiang Zhao\textsuperscript{2}, Shengyu Chen\textsuperscript{3}\\
Yanchi Liu\textsuperscript{3}, Wei Cheng\textsuperscript{3}, Xi Zhu\textsuperscript{1}\\
Guoning Zhang\textsuperscript{1}, Dimitris N. Metaxas\textsuperscript{1}, Haifeng Chen\textsuperscript{3}\\[0.6em]
\normalfont\small
\textsuperscript{1}Rutgers University\quad
\textsuperscript{2}Meta\quad
\textsuperscript{3}NEC Labs America
\end{minipage}}

\hypersetup{
  hidelinks,
  pdfauthor={Bangwei Guo, Xujiang Zhao, Shengyu Chen, Yanchi Liu, Wei Cheng, Xi Zhu, Guoning Zhang, Dimitris N. Metaxas, Haifeng Chen},
  pdftitle={Knossos and Ariadne: Benchmarking and Learning Complete Diagram Topology Extraction with Vision--Language Models}
}

\begin{document}
\maketitle
\pagestyle{fancy}
\fancyhead{}
\lhead{Under review}

\begin{abstract}
Structural diagrams are widely used to represent complex systems and relational
information across scientific, engineering, procedural, and spatial domains. Recent vision--language models (VLMs) have become increasingly capable of
recognizing diagram elements and reasoning about their content, while complete
diagram topology extraction remains comparatively underexplored. In this paper, we study \textbf{diagram-to-graph topology extraction}:
extracting all diagram entities and the complete relations among them.
To enable large-scale supervised training and systematic evaluation of this
task, we introduce \textbf{Knossos}, a benchmark of 19,200 diagrams across six diverse domains, with 245,179 nodes
and 439,740 edges. Its symbolic generation process provides exact alignment
between rendered diagrams and annotations of complete topology, relation types,
and connector geometry. To address the modeling challenge of complete topology extraction, we also
present \textbf{Ariadne}, a structured framework that decomposes the task into
node inventory extraction and source-conditioned edge prediction. Extensive experiments show that training on Knossos substantially
improves complete topology extraction in smaller open-source VLMs.
Ariadne further improves over one-step extraction under matched supervision,
demonstrating the additional benefit of structured decomposition. It achieves
the highest average Edge F1 among the evaluated methods on Knossos,
while both backbone variants also improve over their unadapted counterparts
on the real-world external benchmark. Code and benchmark are available at
\url{https://github.com/bangwayne/knossos_Ariadne_Public}.

\end{abstract}

\section{Introduction}

Structural diagrams are essential for representing complex structured and
relational information across science, engineering, education, and technical
domains. Recent advances in large vision--language models (VLMs) have broadened the scope of diagram understanding, spanning scientific and chart
question answering, visual-graph reasoning, and specialized engineering-diagram
analysis~\citep{masry2022chartqa,zhu2025benchmarking,zhao2025can,li2025eee}. These advances have improved the ability of VLMs to recognize
diagram elements and reason about their content. However, structural diagrams
often contain many interacting entities and densely routed relations, making
complete structural understanding substantially more challenging than
recognizing individual elements~\citep{hou2024vision,hu2026diagram2structure}. 

Despite this progress, existing diagram-understanding tasks typically evaluate
only selected entities and relations, for example through question answering
or localized reasoning~\citep{lu2022learn,li2024synthesize}. Such evaluations do not require extracting the complete topology: a model may
answer individual questions correctly while still omitting, reversing, or
hallucinating other relations~\citep{zhu2025benchmarking, hu2026diagram2structure}. We therefore study \textbf{diagram-to-graph
topology extraction}, which aims to extract the complete graph topology encoded in a diagram, where diagram entities correspond to nodes
and their relations to edges. Once the complete topology is available, many structure-centric question
answering and reasoning tasks become simpler by directly analyzing the
extracted textual graph representation.

Complete diagram-to-graph topology extraction, however, faces two key
challenges: the scarcity of large-scale benchmarks that support both training
and evaluation, and the difficulty of modeling complete diagram topology. On the data side, constructing complete topology supervision requires
exhaustively labeling all diagram entities and relations, making large-scale
dataset construction far more demanding than question-answer supervision~\citep{guo2026topoagent, hu2026diagram2structure}. On the modeling side, VLMs must identify all relevant entities and extract their complete relations from potentially dense diagrams. As diagram complexity grows,
the resulting graph serialization becomes longer and harder to generate
consistently, increasing the likelihood of omissions, hallucinations, and
structural inconsistencies~\citep{jiang2026kvsmooth, zheng2025lvlms}.

To address these challenges, we introduce two complementary contributions:
\textbf{Knossos}, a large-scale benchmark for supervised topology learning
and evaluation, and \textbf{Ariadne}, a structured framework for complete topology extraction\footnote{The names draw from the Greek myth of the Labyrinth at
Knossos, where Ariadne gave Theseus a thread to navigate its passages. Here,
Knossos evokes the benchmark's intricate structures, while Ariadne provides
the thread for extracting their underlying topology.}. Knossos is generated
from symbolic topology specifications, ensuring exact alignment between
rendered diagrams and complete topology annotations by construction. This pipeline scales to 19,200 diagrams across six domains while varying
topology structures, layouts, connector routes, and visual assets. Ariadne
addresses the modeling challenge by first extracting a complete node inventory
and then decomposing topology extraction into source-group edge predictions,
replacing a single long graph serialization with shorter structured generations
that are deterministically assembled into the complete topology. We train Ariadne on the Knossos training split using its complete topology
annotations as supervision. Extensive experiments show that Knossos provides
effective large-scale supervision for complete topology extraction, while Ariadne further improves over matched one-step models
through structured source-conditioned prediction. Evaluation on an external real-world benchmark further demonstrates
cross-domain generalization beyond the synthetic training distribution. Our contributions are summarized as follows:

\begin{itemize}
    \item We establish a large-scale supervised setting for
    diagram-to-graph topology extraction, moving beyond isolated
    entities or relations toward complete diagram topology extraction with
    vision--language models.

    \item We introduce Knossos, a large-scale benchmark containing
    19,200 diagrams across six diverse domains, with complete annotations of
    nodes, directed and typed relations, and connector geometry for both
    training and evaluation.

    \item We present Ariadne, a structured framework that decomposes
    complete topology extraction into node inventory construction and short
    source-conditioned edge predictions.

    \item We conduct extensive experiments across Knossos and an external real-world benchmark, demonstrating the effectiveness of large-scale topology supervision, the benefit of structured decomposition, and cross-domain generalization.
\end{itemize}

\section{Task Formulation and the \textsc{Knossos} Benchmark}
\label{sec:benchmark}

\begin{figure}[t]
    \centering
    \includegraphics[width=\linewidth]{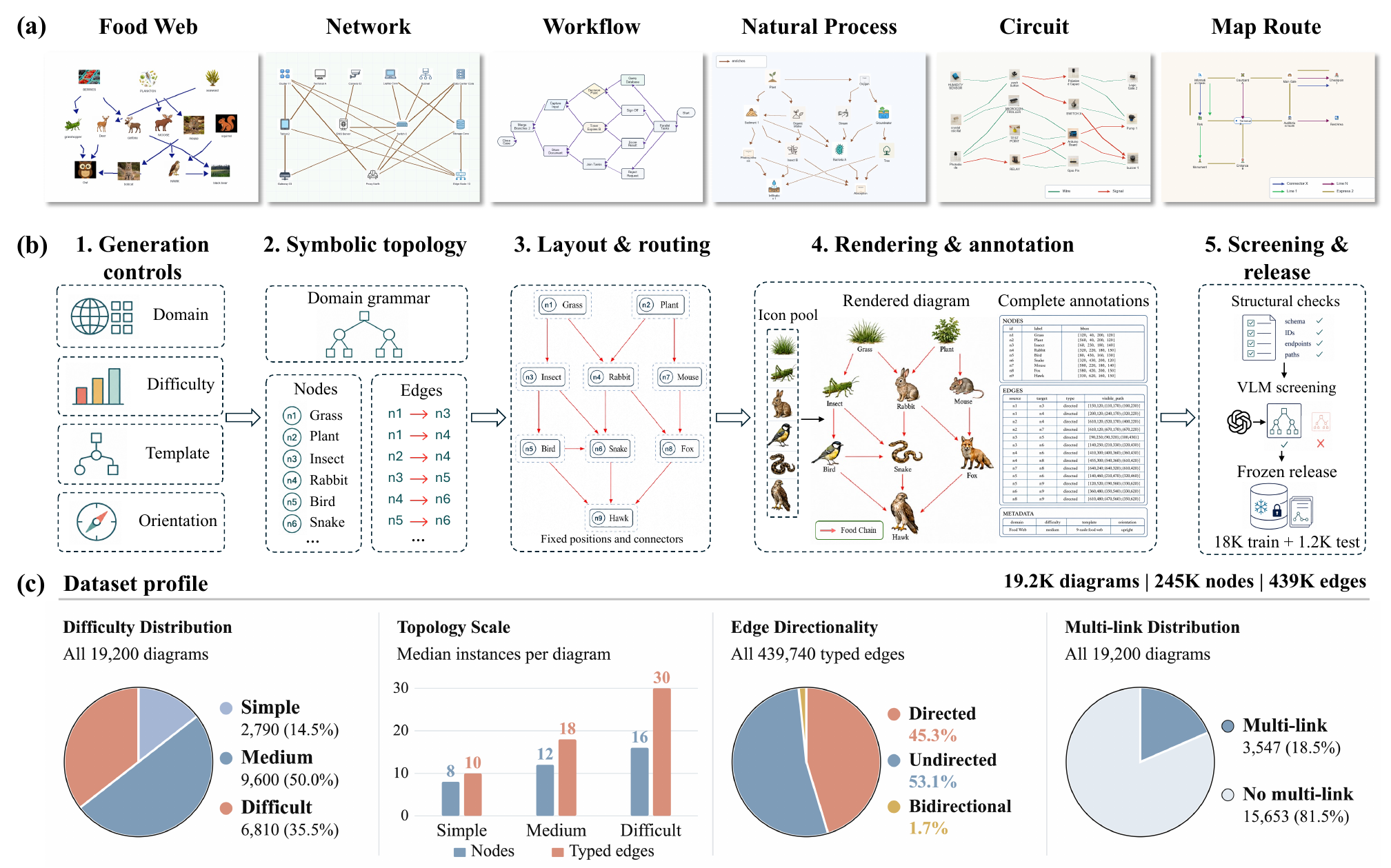}
    \caption{
    \textbf{Overview of the Knossos benchmark.}
    (a) Representative diagrams from the six domains.
    (b) The generation pipeline samples generation controls, instantiates a symbolic topology, places nodes and routes edges, and renders the diagram together with aligned node, typed-edge, and geometry annotations. Structural validation and VLM-based screening produce the frozen training and test release.
    (c) Dataset statistics summarize difficulty, topology scale, edge directionality, and multi-link prevalence across 19,200 diagrams, 245,179 nodes, and 439,740 typed edges.
    }
    \label{fig:knossos_overview}
    \vspace{-0.5\baselineskip}
\end{figure}

\subsection{Diagram-to-Graph Topology Extraction}
\label{sec:task_definition}

Given a diagram image $I$, the task is to extract its topology
$G=(V,E)$, where $V$ is the set of diagram nodes and $E$ is a multiset of
directed edges. Each node $v_i\in V$ has a diagram-local identifier and a
visible text label, and each edge $e=(s,t)\in E$ connects source node $s$ to
target node $t$. When relation semantics are defined, an edge is additionally
assigned a type $r$, yielding a typed edge $(s,t,r)$. Multiple typed edges may
share the same endpoint pair, allowing distinct relations between the same two
nodes to be preserved.

\subsection{Knossos: Multi-Domain Synthetic Diagram Benchmark}

We introduce
Knossos, a large-scale synthetic diagram benchmark for supervised topology
extraction. Knossos contains 19,200 diagrams spanning six
complementary domains: Food Web, Network, Workflow, Natural Process, Circuit,
and Map Route, with representative examples shown in
Figure~\ref{fig:knossos_overview}(a). Together, these domains cover scientific, engineering,
procedural, and spatial structures while introducing diverse extraction
challenges, including dense directed connectivity, unarrowed connections,
branching and merging structures, legend-defined relation types, and multiple
typed edges between common endpoint pairs. The dataset contains 245,179 node
instances and 439,740 edge instances, with the nodes drawn from 310
domain-specific canonical concepts. The frozen release is divided into 18,000
training images and 1,200 held-out test images. Each diagram is paired with
its complete graph topology, relation types when defined by the diagram, and
geometric annotations for both nodes and edges, including node bounding boxes
and connector paths. These annotations support both topology extraction and
geometric grounding.

\subsection{Symbolic Generation Pipeline}

Figure~\ref{fig:knossos_overview}(b) summarizes the Knossos generation
pipeline. The generator samples the domain, difficulty, structural template,
and orientation, instantiates nodes from the corresponding domain vocabulary,
and constructs edges using the domain-specific grammar and selected template.
Once the topology is fixed, the layout stage assigns node positions and
bounding boxes, after which connectors are routed between the corresponding
node boundaries. Depending on the domain and template, connectors may follow
direct, bent, or orthogonal paths, and the routed connector paths are retained
as geometric traces associated with the corresponding edges. Finally, the
renderer combines the fixed topology, layout, connector geometry, and visual
assets to produce the diagram image. Complete annotations are serialized from
the same finalized node and edge objects, including node labels and bounding
boxes, graph topology, relation types when defined, and connector traces. This
shared representation keeps the rendered diagram and its annotations aligned
by construction. Detailed domain grammars and rendering configurations are
provided in Appendix~\ref{app:benchmark_details}.



\subsection{Structural and Visual Diversity}

Structural diversity is introduced through parameterized, domain-specific
template families rather than unconstrained graph sampling. These templates
cover hierarchical, clustered, branching, layered, and spatial structures.
Seeded rules vary graph size, density, depth, branching, cross-links, relation
types, and parallel edges within each grammar. Difficulty controls both
topology and presentation: harder samples generally contain more nodes and
edges, more layers and cross-links, and longer, more intersecting connector
paths. Training and test instances are generated with disjoint random seeds
for template instantiation, topology sampling, and layout variation. 

Visual diversity is introduced through offline-prepared, label-matched asset
pools, with each eligible canonical concept associated with multiple visual
styles and variants; nodes without image assets are rendered programmatically
using role-specific shapes. At rendering time, assets are sampled only from
the pool corresponding to the selected node concept. Seeded presentation
choices further vary, where applicable, orientation, label casing, background,
connector appearance, arrowhead size, and legend placement. Visible labels are
drawn by the renderer rather than by the asset-generation model, so visual
variation does not alter the underlying node identity or graph topology.
Implementation details are provided in
Appendix~\ref{app:asset_pipeline}.

\subsection{Quality Control and Dataset Release}

As summarized in Figure~\ref{fig:knossos_overview}(b), each generated sample
undergoes structural validation followed by VLM-based visual screening.
Structural checks focus on basic annotation and graph consistency, including
format validity, completeness, and duplicate detection. A separate VLM-based screening step filters out samples with clearly
unreadable text or visually unclear diagram elements. The final release is constructed by stratified selection across domains and
difficulty levels, replacing failed samples with valid candidates from the
same stratum. The frozen benchmark contains 18,000 training images and 1,200
held-out test images, with 3,000 training and 200 test examples per domain.
Figure~\ref{fig:knossos_overview}(c) summarizes the resulting benchmark
statistics. Additional screening and release details are provided in
Appendix~\ref{app:qc_release}.


\section{Ariadne}
\label{sec:method}

\begin{figure}[t]
    \centering
    \includegraphics[width=\linewidth]{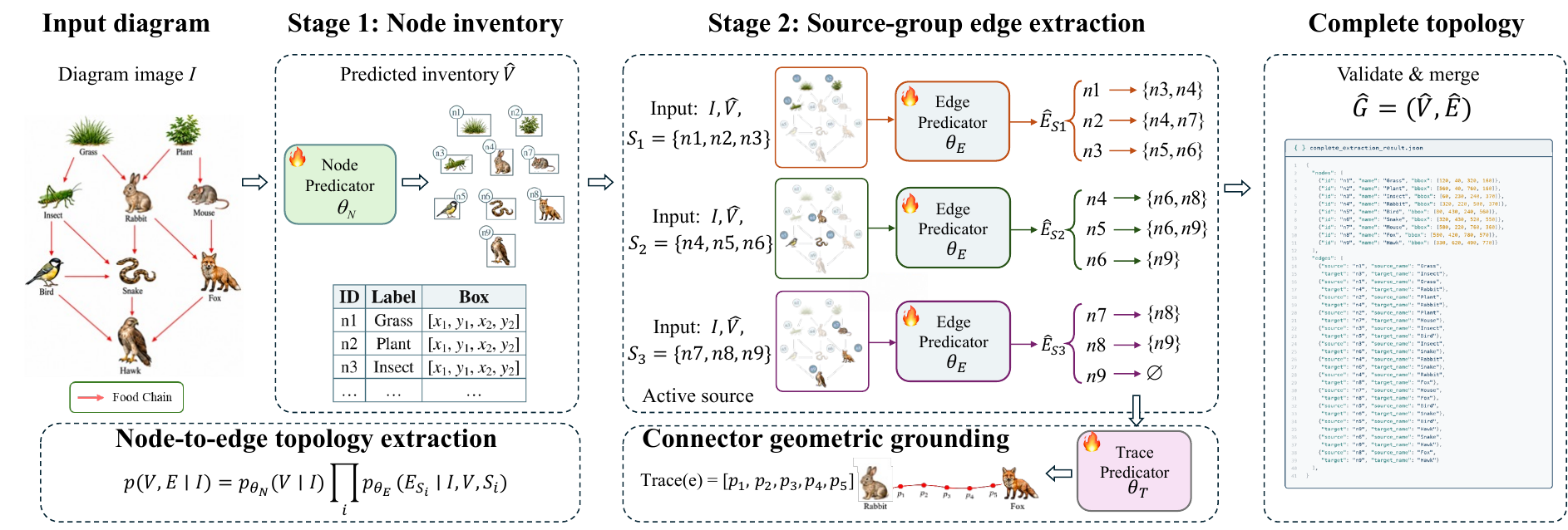}
    \caption{\textbf{Overview of Ariadne.}
    Ariadne decomposes topology extraction into node inventory prediction
    and source-group edge extraction. Edge prediction is restricted to
    active sources within each group while retaining all predicted nodes
    as candidate targets. The group-wise edge multisets are validated and
    merged into the final graph, with node bounding boxes and separately
    predicted connector traces providing auxiliary geometric grounding.}
    \label{fig:ariadne_overview}
\end{figure}

\subsection{Node-to-Edge Topology Extraction}

Figure~\ref{fig:ariadne_overview} illustrates the overall framework. Following the topology formulation in
Section~\ref{sec:task_definition}, Ariadne decomposes topology extraction into
node inventory prediction followed by source-conditioned edge prediction. The predicted
node inventory provides a fixed reference vocabulary and is partitioned into
$K$ source groups $\mathcal{S}(V)=\{S_1,\ldots,S_K\}$ with target group size
$g$. For each $S_i$, the edge predictor outputs only the typed edges
originating from nodes in $S_i$. The group-wise edge multisets are then merged
deterministically to form the complete graph.

We model this decomposition as
\begin{equation}
\label{eq:topoagent_factorization}
p_{\theta}(V,E\mid I)
=
p_{\theta_N}(V\mid I)
\prod_{i=1}^{K}
p_{\theta_E}(E_{S_i}\mid I,V,S_i),
\end{equation}
where $E_{S_i}$ is the multiset of directed edges originating from sources in
$S_i$. At inference, the predicted group-wise edge multisets are merged as
\begin{equation}
\hat E=\biguplus_{i=1}^{K}\hat E_{S_i},
\qquad
\hat G=(\hat V,\hat E).
\end{equation}

This decomposition replaces a single long graph serialization with shorter,
source-conditioned generations and, under our serialization setting, removes
cross-group ordering ambiguity. Under a position-sensitive decoding assumption,
the shorter generation horizon may also reduce error accumulation at later
output positions. We formalize these properties and their assumptions in
Appendix~\ref{app:decomposition}.

\subsection{Two-Stage Topology Prediction}

\paragraph{Stage 1: Node inventory.}
The node predictor first identifies the diagram nodes and their visible text
labels, producing the predicted node inventory $\hat V$. Each node is localized
with a bounding box $b_v=[x_1,y_1,x_2,y_2]$ for geometric grounding, while a
diagram-local identifier is included only to provide a stable reference in
later stages. The output contains only node records and no edge or connector
predictions. At test time, the predicted node inventory, rather than the
ground-truth inventory, is passed to the next stage.

\paragraph{Stage 2: Source-group edge extraction.}
Given $\hat V$, Ariadne processes the source groups
$\{S_i\}_{i=1}^{K}$ independently. Each query contains the diagram, the
complete predicted node inventory, and the active sources in $S_i$. The edge predictor may emit relations only from these active sources, while any node
in $\hat V$ remains available as a target. For each active source, it emits one
\texttt{CHECK\_NODE} block:
\begin{ariadneoutput}{Example Stage 2 output for one active source}
<CHECK_NODE id="n4" name="Sensor">
  <EDGE source="n4" source_name="Sensor"
        target="n6" target_name="Controller"
        type="signal"/>
  <EDGE source="n4" source_name="Sensor"
        target="n8" target_name="Ground"
        type="ground"/>
</CHECK_NODE>
\end{ariadneoutput}

Relation types are predicted only when defined by the diagram legend; otherwise
the type is left empty. Directionality is encoded by the ordered
source--target pair, with undirected or bidirectional connectors represented by
reciprocal edges. Multiple typed relations between the same endpoints remain
distinct multiset elements, and an empty \texttt{CHECK\_NODE} block denotes a
source with no outgoing edge. Finally, all source-group predictions are validated and merged into $\hat E$.

\subsection{Geometric Grounding}

Beyond symbolic topology, Ariadne associates each graph primitive with
explicit geometric evidence. Nodes are localized by the bounding boxes
predicted in Stage~1, while each predicted edge is grounded by an ordered
connector trace generated after the Stage~2 edge set is fixed. For each edge,
the trace predictor outputs five points sampled along its visible connector
from source to target. Training targets are obtained by uniformly sampling the
final rendered connector path by arc length, so the endpoints localize the
connector boundaries and the intermediate points capture its routed geometry.
Reciprocal edges use the same connector path in reverse order. The trace serves
only as auxiliary geometric grounding and does not modify the predicted graph
topology. Detailed results are provided in
Appendix~\ref{app:geometry_comparison}.

\section{Experiments}

\subsection{Experimental Setup}
\label{sec:experimental_setup}

\paragraph{Implementation and evaluation protocol.}
We instantiate Ariadne with GLM-4.6V-Flash~\citep{hong2025glm} and Qwen3-VL-8B~\citep{bai2025qwen3} as two
vision--language backbones, each using separate adapters for the node, edge,
and trace predictors. All main
Knossos results are evaluated on the frozen 1,200-image held-out test set,
with 200 images from each of the six domains. At inference, the predicted node inventory is partitioned
into source groups of target size 3 for edge extraction, while the trace
predictor is applied only after the symbolic edge set is fixed. Complete
training and decoding details are provided in
Appendix~\ref{app:training_configuration}, with prompt templates in
Appendix~\ref{app:training_records}. We use macro-averaged Node F1 and Edge F1 scores as the primary topology metrics.
Node F1 measures node identification after matching predictions to reference
nodes, while Edge F1 evaluates directed source--target connectivity between
matched nodes. Scores are averaged across
images within each domain, and the overall Knossos score is the mean across the
six domains. Additional relation-type and connector-grounding metrics are
reported in ablation analyses and appendix. We additionally evaluate generalization on a 162-image subset of the external
TopoBench-180 test set~\citep{guo2026topoagent}, with preprocessing details
provided in Appendix~\ref{app:topobench_preprocessing}.
 We report only Node F1 and Edge F1 on this benchmark.

\paragraph{Baselines.}
We organize the baselines into three groups.
Open-source VLMs include Qwen3-VL-32B and
Qwen3-VL-8B~\citep{bai2025qwen3},
GLM-4.6V-Flash~\citep{hong2025glm},
MiMo-VL-7B-RL~\citep{yue2025mimo},
Keye-VL-1.5-8B~\citep{yang2025kwai},
InternVL3.5-8B~\citep{wang2025internvl3}, and
Ovis2.5-9B~\citep{lu2025ovis2}.
Closed-source VLMs include
GPT-4o~\citep{hurst2024gpt},
GPT-5.6~\citep{singh2025openai}, and
Claude Opus 5~\citep{anthropic2026opus5}.
For all open- and closed-source VLM baselines, we use direct full-graph
generation, requiring the model to produce all diagram nodes and edges in a
single response. This setting provides a direct comparison with Ariadne's
decomposed node-to-edge prediction strategy. Structured visual-reasoning methods include
Pixel Reasoner~\citep{su2026pixel},
Chain of Region~\citep{li2025chain},
Speculative Verdict~\citep{liu2026small}, and
TopoAgent~\citep{guo2026topoagent}. We instantiate Ariadne with two backbone configurations,
GLM-4.6V-Flash (9B) and Qwen3-VL-8B, allowing us to evaluate the proposed
decomposition across two different open-source VLM families.

\subsection{Node extraction}

\begin{figure*}[t]
    \centering
    \includegraphics[width=\textwidth]{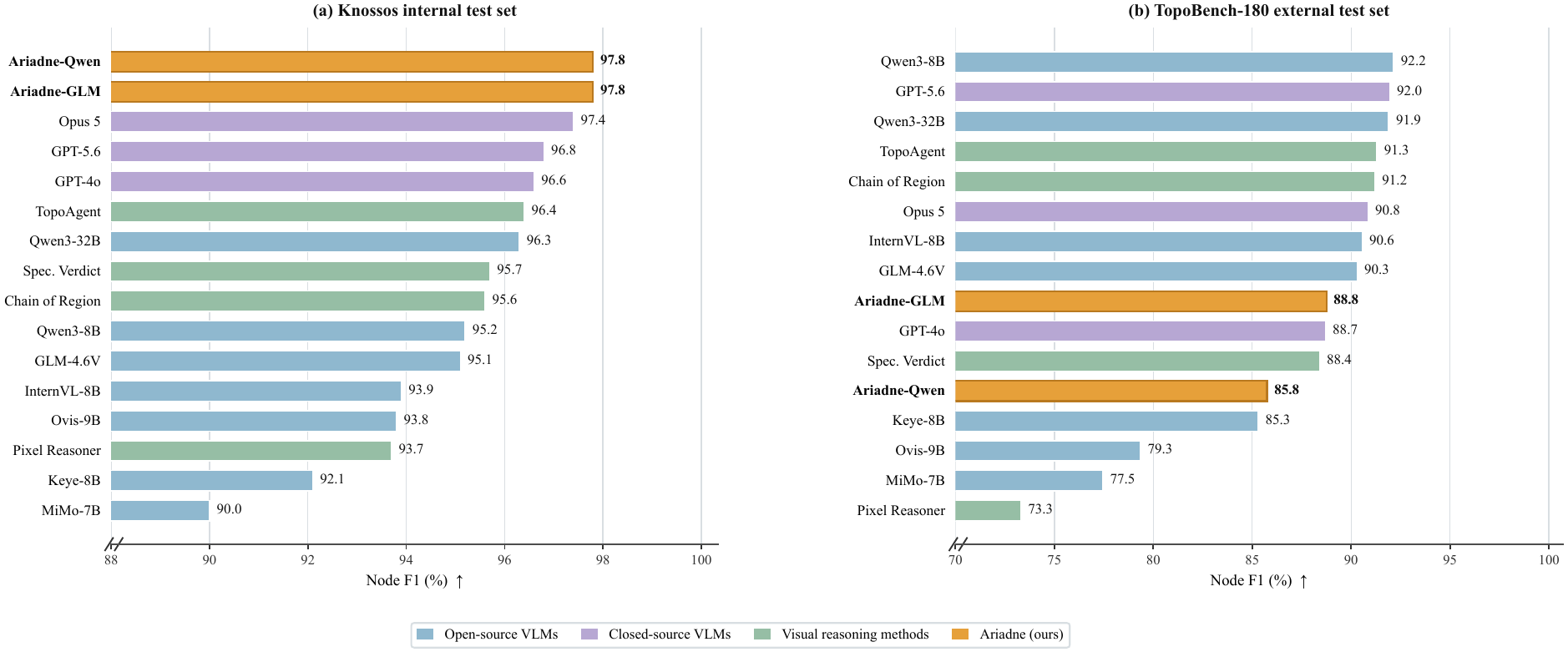}
    \caption{Node extraction results on the internal and external test
    sets. Panel (a) reports the unweighted mean Node F1 across the six
    Knossos domains. Panel (b) reports the unweighted mean across the
    Web-style and Network-style subsets of TopoBench-180. Colors distinguish
    open-source VLMs, closed-source VLMs, visual reasoning methods, and our
    Ariadne models. Full model names are listed in Table~\ref{tab:main}.}
    \label{fig:node_f1_internal_external}
\end{figure*}

Figure~\ref{fig:node_f1_internal_external} compares node extraction results on the
internal and external test sets. Node extraction is highly reliable on the
Knossos internal test set: Ariadne achieves a macro-average Node F1 of 97.8\%
with both the GLM- and Qwen3-based variants, while strong open-source VLMs also
generally exceed 95\%. Recent closed-source VLMs are already close
to this ceiling, with Opus~5 reaching 97.4\% and GPT-5.6 reaching 96.8\%.
Together, these results suggest that current multimodal models already provide
strong OCR and visual-semantic recognition for identifying diagram entities.
Performance decreases on TopoBench-180, suggesting that additional
node-specific training may introduce some distribution specialization. Given
the already strong node recognition capabilities of modern VLMs, a separately
trained node predictor may therefore not always be necessary. This further
motivates focusing model capacity and evaluation on relation and topology extraction, where the performance gap remains substantially larger. Detailed
domain-level results and additional analysis are provided in
Appendix~\ref{app:node_results}.

\subsection{Edge extraction}

Table~\ref{tab:main} compares Ariadne with a broad set of baselines on
Edge F1. On Knossos, recent closed-source VLMs already show strong direct
topology-extraction capability: GPT-5.6 reaches 81.1\% average Edge F1,
substantially outperforming all direct open-source VLMs. Ariadne improves
further, reaching 91.6\% with the Qwen3-based variant and 88.8\% with the
GLM-based variant. Ariadne-Qwen uses an 8B backbone yet substantially exceeds
the 32B Qwen3-VL model under direct extraction (91.6\% vs.\ 54.7\%). Relative
to the corresponding direct backbones, Ariadne improves Qwen3-VL-8B from
46.1\% to 91.6\% and GLM-4.6V-Flash from 38.9\% to 88.8\%, reflecting the
combined effects of task adaptation and the proposed structured decomposition.

On TopoBench-180, closed-source VLMs show the strongest generalization, with
Claude Opus~5 reaching 81.2\% on Web-style diagrams and GPT-5.6 reaching
68.7\% on Network-style diagrams. This suggests that broad multimodal
pretraining provides strong out-of-distribution topology understanding without
task-specific adaptation. Ariadne nevertheless substantially improves over
its corresponding backbones: the GLM-based variant increases from 55.2\% to
73.4\% on Web-style diagrams and from 45.1\% to 55.6\% on Network-style
diagrams, while the Qwen3-based variant improves from 44.1\% to 68.6\% and
from 49.8\% to 53.2\%, respectively. The GLM-based variant also generalizes
better than the Qwen3-based variant despite its lower in-domain score,
highlighting the distinction between in-distribution performance and external
generalization.

\begin{wrapfigure}{r}{0.5\textwidth}
    \vspace{-1.0\baselineskip}
    \centering
    \includegraphics[width=\linewidth]{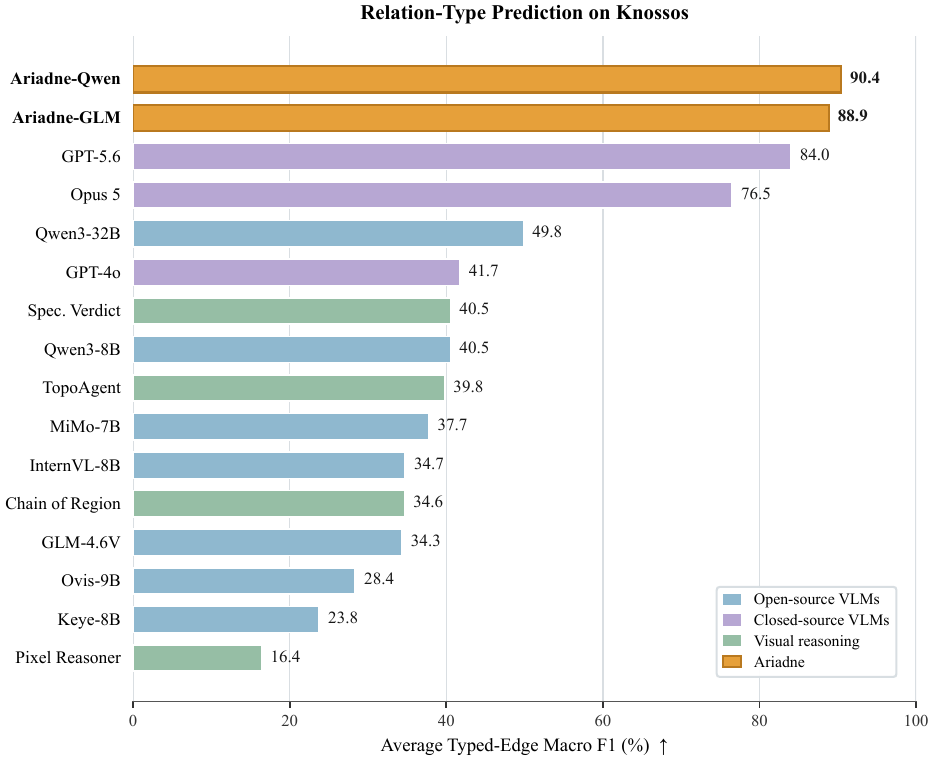}
    \caption{Average Typed-Edge F1 scores across Natural Process, Circuit,
    and Map Route. Each bar is the arithmetic mean of the three domain-level
    Macro F1 scores reported in Appendix Table~\ref{tab:type_f1}.}
    \label{fig:typed_edge_macro_average}
    \vspace{-0.6\baselineskip}
\end{wrapfigure}

Edge F1 evaluates whether the two endpoints are correctly recovered, but
does not require the predicted relation type to be correct. Typed-Edge F1 is
stricter: a prediction is counted as correct only when both endpoints match
and the edge type indicated by the diagram legend is also extracted correctly.
Figure~\ref{fig:typed_edge_macro_average} summarizes relation-type prediction
by averaging Typed-Edge F1 over Natural Process, Circuit, and Map Route,
the three domains with explicit relation semantics. Ariadne-Qwen ranks first
at 90.4\%, followed by Ariadne-GLM at 88.9\%. Both outperform the strongest
baseline, GPT-5.6 at 84.0\%, by 6.4 and 4.9 points, respectively. The gap is
substantially larger relative to direct open-source VLMs: the strongest of
these, Qwen3-VL-32B, reaches 49.8\%. This aggregate comparison indicates that
Ariadne's gains extend beyond identifying the correct endpoints to extracting
the semantics of their relations. Complete domain-level results are provided
in Appendix Table~\ref{tab:type_f1}.

\subsection{Analysis and Ablations}
\label{sec:analysis_ablations}

\paragraph{When does staged topology extraction help?}

Table~\ref{tab:node_edge_decomposition} compares one-step and staged topology
extraction across models with different capacities and training settings.
Staging does not improve GPT-5.6 and instead reduces its Edge F1, suggesting
that frontier models may already have sufficient capacity for joint entity and
relation extraction. In contrast, decomposition provides a modest gain for the
base Qwen3-VL-8B model and a much larger gain after task-specific adaptation:
Ariadne-Qwen improves from 80.0\% to 91.6\% Edge F1. Node F1 changes only
marginally, indicating that the improvement primarily comes from more reliable
edge extraction. Overall, our results suggest that staged decomposition is
particularly beneficial for smaller task-adapted models, while larger frontier
models benefit less from explicitly separating the two steps.

\begin{table*}[t]
\caption{Edge extraction results across six Knossos domains and two
external TopoBench-180 domains, measured by Edge F1 scores (\%).}
\label{tab:main}
\centering
\small
\resizebox{\textwidth}{!}{%
\begin{tabular}{lccccccccc}
\toprule
& \multicolumn{7}{c}{Knossos}
& \multicolumn{2}{c}{TopoBench-180} \\
\cmidrule(lr){2-8}
\cmidrule(lr){9-10}
Method
& Food Web & Network & Workflow
& Natural Process & Circuit & Map Route & Avg.
& Web-style & Network-style \\
\midrule

\multicolumn{10}{l}{\textit{Open-source VLMs}} \\
\cmidrule(lr){1-10}
Qwen3-VL-32B
& 28.8 & 48.7 & 76.4 & 65.8 & 36.7 & 71.9 & 54.7 & 54.7 & 52.3 \\
Qwen3-VL-8B
& 27.5 & 47.0 & 59.9 & 49.5 & 31.6 & 61.1 & 46.1 & 44.1 & 49.8 \\
GLM-4.6V-Flash
& 30.0 & 25.0 & 46.3 & 48.4 & 27.3 & 56.3 & 38.9 & 55.2 & 45.1 \\
MiMo-VL-7B-RL
& 24.7 & 37.1 & 54.3 & 40.2 & 26.1 & 61.7 & 40.7 & 42.2 & 46.5 \\
Keye-VL-1.5-8B
& 30.7 & 36.7 & 51.1 & 44.3 & 22.0 & 49.2 & 39.0 & 47.9 & 45.5 \\
InternVL3.5-8B
& 22.8 & 46.0 & 52.5 & 44.8 & 28.0 & 50.3 & 40.7 & 41.9 & 46.2 \\
Ovis2.5-9B
& 25.8 & 37.5 & 56.7 & 39.0 & 25.2 & 45.8 & 38.3 & 42.1 & 48.8 \\
\cmidrule(lr){1-10}
\addlinespace[3pt]
\multicolumn{10}{l}{\textit{Closed-source VLMs}} \\
\cmidrule(lr){1-10}
GPT-4o
& 16.8 & 43.3 & 46.8 & 43.1 & 30.5 & 67.0 & 41.2 & 32.8 & 54.1 \\
GPT-5.6
& 61.9 & 91.0 & 80.5 & 89.6 & 77.0 & 86.6 & 81.1 & 80.3 & \textbf{68.7} \\
Opus 5
& 54.7 & 51.1 & 85.6 & 96.5 & 56.9 & 77.5 & 70.4 & \textbf{81.2} & 51.8 \\
\cmidrule(lr){1-10}
\addlinespace[3pt]
\multicolumn{10}{l}{\textit{Visual reasoning methods}} \\
\cmidrule(lr){1-10}
Pixel Reasoner
& 23.0 & 31.4 & 35.4 & 31.2 & 24.3 & 45.6 & 31.8 & 32.6 & 35.3 \\
Chain of Region
& 18.2 & 40.6 & 45.6 & 40.2 & 28.3 & 57.1 & 38.3 & 35.0 & 51.8 \\
Speculative Verdict
& 23.4 & 36.8 & 64.2 & 50.9 & 28.9 & 59.4 & 43.9 & 42.8 & 47.4 \\
TopoAgent
& 28.4 & 51.6 & 54.5 & 41.4 & 37.6 & 65.5 & 46.5 & 45.8 & 56.8 \\
\cmidrule(lr){1-10}
\addlinespace[3pt]
\multicolumn{10}{l}{\textit{Ours}} \\
\cmidrule(lr){1-10}
\textbf{Ariadne (GLM-based)}
& 80.4 & 88.5 & 94.3 & 96.7 & 84.3 & 88.3 & 88.8 & 73.4 & 55.6 \\
\textbf{Ariadne (Qwen3-based)}
& \textbf{87.2} & \textbf{93.5} & \textbf{95.5} & \textbf{97.8} & \textbf{86.4} & \textbf{89.3} & \textbf{91.6} & 68.6 & 53.2 \\

\bottomrule
\end{tabular}%
}
\end{table*}
\FloatBarrier

\Needspace{0.55\textheight}




\paragraph{What limits Ariadne's external generalization?}
Table~\ref{tab:node_source_ablation} examines why the large gains observed on
the internal test set generalize only partially to the external test set. We
keep the trained Qwen3 edge predictor fixed and vary only the upstream node
inventory. Replacing Ariadne's task-adapted node predictions with those from
the base Qwen3-VL-8B raises Node F1 from 85.8\% to 92.2\% and Edge F1 from
60.9\% to 67.3\%. This indicates that the more limited external improvement
is driven in substantial part by weaker node generalization under distribution
shift, which constrains the downstream edge predictor. Meanwhile,
general-purpose VLMs already show strong node-recognition generalization on
the external data, suggesting that additional node-specific adaptation may be
less necessary in this setting. With oracle nodes, the frozen edge predictor
reaches 78.2\% macro-F1, 10.9 points above the strongest predicted-node source,
confirming substantial error propagation from node extraction while also
revealing remaining limitations in edge reasoning even when node uncertainty
is removed.

\begin{table*}[t]
\centering
\small

\begin{subtable}[t]{0.46\textwidth}
\centering
\caption{One-step vs.\ staged topology extraction on the Knossos test set.
Node and Edge F1 are macro-averaged across the six domains (\%).}
\label{tab:node_edge_decomposition}
\resizebox{\linewidth}{!}{%
\begin{tabular}{llcc}
\toprule
Model & Protocol & Node F1 $\uparrow$ & Edge F1 $\uparrow$ \\
\midrule
\multirow{2}{*}{GPT-5.6}
  & One-step          & 96.8 & 81.1 \\
  & Staged extraction & 97.0 & 77.1 \\
\addlinespace
\multirow{2}{*}{Qwen3-VL-8B (base)}
  & One-step          & 95.2 & 46.1 \\
  & Staged extraction & 95.1 & 49.3 \\
\addlinespace
\multirow{2}{*}{Ariadne-Qwen}
  & One-step          & 97.1 & 80.0 \\
  & Staged extraction & 97.8 & 91.6 \\
\bottomrule
\end{tabular}%
}
\end{subtable}
\hfill
\begin{subtable}[t]{0.52\textwidth}
\centering
\caption{Node-source ablation on the external TopoBench test set with a fixed
trained Qwen3 edge predictor.}
\label{tab:node_source_ablation}
\renewcommand{\arraystretch}{1.12}
\resizebox{\linewidth}{!}{%
\begin{tabular}{lcccc}
\toprule
Node source
& \shortstack{Average\\Node F1 $\uparrow$}
& \shortstack{Average\\Edge F1 $\uparrow$}
& \shortstack{Web-style\\Edge F1 $\uparrow$}
& \shortstack{Network-style\\Edge F1 $\uparrow$} \\
\midrule
Ariadne node predictor
& 85.8 & 60.9 & 68.6 & 53.2 \\
Qwen3-VL-8B (base)
& 92.2 & 67.3 & 71.7 & 62.9 \\
GPT-5.6
& 92.0 & 67.1 & 75.0 & 59.2 \\
Oracle nodes
& 100.0 & 78.2 & 78.3 & 78.1 \\
\bottomrule
\end{tabular}%
}
\end{subtable}

\caption{Ablations on staged topology extraction and external generalization.
(a) compares one-step and staged extraction across model and training settings.
(b) varies the upstream node inventory while keeping the trained Qwen3 edge
predictor fixed on the public TopoBench subset.}
\label{tab:topology_ablations}
\end{table*}

\paragraph{What diagram structures remain challenging?}
Table~\ref{tab:structural_slices} examines how diagram structure affects
topology prediction. Performance decreases consistently as diagram difficulty
increases, showing that overall diagram complexity remains a primary source of
error. Harder diagrams typically contain denser connectivity, more cross-links,
and more intersecting connector paths, increasing the difficulty of assigning
relations to the correct endpoints. Multi-link diagrams are also more
challenging, likely because multiple relations must be visually disentangled
and preserved separately rather than collapsed into a single connection. In
contrast, performance remains relatively stable across directed,
bidirectional, and mixed settings, suggesting that relation direction itself
is no longer a major bottleneck for current models. Overall, topology
extraction is affected more by diagram complexity and visually entangled
connectivity than by directionality.

\Needspace{0.6\textheight}
\section{Discussion and Conclusion}


\begin{wraptable}{r}{0.5\textwidth}
    \vspace{-0.8\baselineskip}
    \centering
    \small
    \caption{Edge F1 (\%) across structural subsets. $N$ denotes the number of
    images. To better isolate each structural factor while reducing variation from
    other domain-specific properties, difficulty is averaged across all six Knossos
    domains, directionality is evaluated within Map Route, and multi-link effects
    are evaluated across Circuit, Map Route, and Natural Process.}
    \label{tab:structural_slices}

    \setlength{\tabcolsep}{3.0pt}
    \renewcommand{\arraystretch}{1.05}

    \resizebox{\linewidth}{!}{%
    \begin{tabular}{llrrr}
    \toprule
    Factor & Subset & $N$ & GPT-5.6 & Ariadne-Qwen \\
    \midrule
    \multirow{3}{*}{Difficulty}
      & Simple        & 164 & 94.4 & 96.5\\
      & Medium        & 571 & 84.9 & 94.7 \\
      & Difficult     & 465 & 76.8 & 89.0 \\
    \addlinespace
    \multirow{3}{*}{Direction}
      & Directed      & 132 & 86.7 & 89.0 \\
      & Bidirectional & 37  & 86.7 & 88.6 \\
      & Mixed         & 31  & 86.2 & 91.0 \\
    \addlinespace
    \multirow{2}{*}{Multi-link}
      & Absent        & 384 & 86.0 & 92.0 \\
      & Present       & 216 & 82.1 & 89.6 \\
    \bottomrule
    \end{tabular}%
    }

\end{wraptable}

We studied complete diagram-to-graph topology extraction, where a model must
extract both the visible entities and the directed relations connecting them in the diagram.
To support this task, we introduced Knossos, a benchmark of 19,200 diagrams
across six domains with aligned node, edge, relation-type, and connector
annotations, and Ariadne, which separates node inventory prediction from
source-conditioned edge extraction. On the 1,200-image Knossos test set,
Ariadne-Qwen reaches 91.6\% Edge F1, compared with 81.1\% for GPT-5.6, with
the margin increasing from 2.0 points on simple diagrams to 12.2 points on
difficult diagrams. On the external TopoBench-180 test set, Ariadne substantially improves over
its direct open-source backbones, although the strongest closed-source VLMs
remain more competitive.

\paragraph{Limitations and outlook.}
First, Knossos remains synthetic, and its assets, layouts, connector styles,
and rendering patterns cannot capture the full diversity of real-world
diagrams. Results on the external TopoBench-180 test set show generalization
beyond the synthetic distribution, but a synthetic-to-real gap remains.
Second, Knossos relies on finite domain-specific vocabularies and predefined
graph grammars, limiting the semantic and structural diversity seen during
training. Generalization to unseen entity types, relation semantics, and
topology patterns therefore remains open. Third, Ariadne is sequential: node
errors can propagate to edge extraction, while source grouping requires
multiple model calls per image. Finally, structured decomposition benefits
smaller task-adapted VLMs most, whereas recent closed-source models already
show strong direct topology-extraction capability. Whether these gains persist
with stronger and larger backbones remains to be studied. Broader real-world
evaluation, more efficient decoding, and stronger open-source backbones are
important directions for future work.

\paragraph{Conclusion.}
Our results show that complete diagram topology extraction can be substantially
improved through large-scale supervision and structured prediction. Knossos
provides a large-scale benchmark for training and controlled evaluation, while
Ariadne demonstrates that source-conditioned decomposition can effectively
improve topology extraction, particularly for smaller task-adapted VLMs.
Together, they provide a practical framework for studying diagram-to-graph
topology extraction. We hope Knossos and Ariadne support further progress
toward reliable topology extraction across increasingly diverse real-world
diagrams.

\subsection*{AI Use Statement}

Generative AI and agentic AI tools were used to assist with code implementation,
debugging, code execution, and experiment workflows, as well as synthetic visual
asset generation, the creation and editing of figures and other visual elements,
and language editing of the manuscript. AI agents were also used to execute and
inspect computational workflows under author supervision. All AI-assisted code,
experimental outputs, data-processing results, figures, and text were reviewed
and verified by the authors. The authors take full responsibility for the final
code, data, experiments, analyses, claims, figures, and manuscript.

\subsection*{Reproducibility Statement}
The benchmark construction rules, asset-generation and sampling policies,
annotation schema, validation procedure, training stages, and evaluation
metrics are documented in the main paper and appendix. The complete benchmark,
including 18,000 training images and 1,200 test images, annotations, and visual
assets, is publicly available at
\url{https://huggingface.co/datasets/WayneGuo0011/Knossos}.
Generation code, training recipes, and evaluation scripts are available at
\url{https://github.com/bangwayne/knossos_Ariadne_Public}.

\bibliography{iclr2027_conference}

@inproceedings{shi2026flowgen,
  title={FlowGen: Synthesizing Diverse Flowcharts to Enhance and Benchmark MLLM Reasoning},
  author={Shi, Kaiwen and Liu, Sichen and Lin, Ziyue and Guo, Hangrui and Cheng, Gong},
  booktitle={International Conference on Learning Representations},
  volume={2026},
  pages={19761--19791},
  year={2026}
}

@inproceedings{pan2024flowlearn,
  title={Flowlearn: Evaluating large vision-language models on flowchart understanding},
  author={Pan, Huitong and Zhang, Qi and Caragea, Cornelia and Dragut, Eduard and Jan Latecki, Longin},
  booktitle={ECAI 2024: 27th European Conference on Artificial Intelligence, 19--24 October 2024, Santiago de Compostela, Spain--Including 13th Conference on Prestigious Applications of Intelligent Systems (PAIS 2024)},
  pages={73--80},
  year={2024},
  organization={SAGE Publications Pvt. Ltd 1 Oliver's Yard, 55 City Road, London, EC1Y 1SP}
}

@inproceedings{scholch2022towards,
  title={Towards automatic parsing of structured visual content through the use of synthetic data},
  author={Sch{\"o}lch, Lukas and Steinh{\"a}user, Jonas and Beichter, Maximilian and Seibold, Constantin and Yang, Kailun and Knaeble, Merlin and Schwarz, Thorsten and Maedche, Alexander and Stiefelhagen, Rainer},
  booktitle={2022 26th International Conference on Pattern Recognition (ICPR)},
  pages={1607--1613},
  year={2022},
  organization={IEEE}
}

@inproceedings{kembhavi2016diagram,
  title={A diagram is worth a dozen images},
  author={Kembhavi, Aniruddha and Salvato, Mike and Kolve, Eric and Seo, Minjoon and Hajishirzi, Hannaneh and Farhadi, Ali},
  booktitle={European conference on computer vision},
  pages={235--251},
  year={2016},
  organization={Springer}
}

@inproceedings{masry2022chartqa,
  title={Chartqa: A benchmark for question answering about charts with visual and logical reasoning},
  author={Masry, Ahmed and Tan, Jia Qing and Joty, Shafiq and Hoque, Enamul and others},
  booktitle={Findings of the association for computational linguistics: ACL 2022},
  pages={2263--2279},
  year={2022}
}

@inproceedings{zhu2025benchmarking,
  title={Benchmarking and improving large vision-language models for fundamental visual graph understanding and reasoning},
  author={Zhu, Yingjie and Bai, Xuefeng and Chen, Kehai and Xiang, Yang and Yu, Jun and Zhang, Min},
  booktitle={Proceedings of the 63rd Annual Meeting of the Association for Computational Linguistics (Volume 1: Long Papers)},
  pages={30678--30701},
  year={2025}
}

@inproceedings{li2025eee,
  title={Eee-bench: A comprehensive multimodal electrical and electronics engineering benchmark},
  author={Li, Ming and Zhong, Jike and Chen, Tianle and Lai, Yuxiang and Psounis, Konstantinos},
  booktitle={2025 IEEE/CVF Conference on Computer Vision and Pattern Recognition (CVPR)},
  pages={13337--13349},
  year={2025},
  organization={IEEE}
}

@inproceedings{zhao2025can,
  title={Can multimodal foundation models understand schematic diagrams? an empirical study on information-seeking qa over scientific papers},
  author={Zhao, Yilun and Wang, Chengye and Li, Chuhan and Cohan, Arman},
  booktitle={Findings of the Association for Computational Linguistics: ACL 2025},
  pages={18598--18631},
  year={2025}
}

@inproceedings{hu2026diagram2structure,
  title={Diagram2Structure: Unlocking LLMs' Diagram Comprehension through DiagramDiff, a Framework for Structuring Offline Diagrams},
  author={Hu, Haoxiang and Li, Yaokun and Huang, Zeyuan and Gao, Cangjun and He, Qiang and Li, Qingkun and Deng, Xiaoming and Ma, Cuixia and Lai, Yu-Kun and Liu, Yong-Jin and others},
  booktitle={Proceedings of the IEEE/CVF Conference on Computer Vision and Pattern Recognition},
  pages={24395--24404},
  year={2026}
}

@article{hou2024vision,
  title={Do vision-language models really understand visual language?},
  author={Hou, Yifan and Giledereli, Buse and Tu, Yilei and Sachan, Mrinmaya},
  journal={arXiv preprint arXiv:2410.00193},
  year={2024}
}

@article{lu2022learn,
  title={Learn to explain: Multimodal reasoning via thought chains for science question answering},
  author={Lu, Pan and Mishra, Swaroop and Xia, Tanglin and Qiu, Liang and Chang, Kai-Wei and Zhu, Song-Chun and Tafjord, Oyvind and Clark, Peter and Kalyan, Ashwin},
  journal={Advances in neural information processing systems},
  volume={35},
  pages={2507--2521},
  year={2022}
}

@inproceedings{li2024synthesize,
  title={Synthesize step-by-step: Tools, templates and LLMs as data generators for reasoning-based chart VQA},
  author={Li, Zhuowan and Jasani, Bhavan and Tang, Peng and Ghadar, Shabnam},
  booktitle={2024 IEEE/CVF Conference on Computer Vision and Pattern Recognition (CVPR)},
  pages={13613--13623},
  year={2024},
  organization={IEEE}
}

@inproceedings{liu2024mmbench,
  title={Mmbench: Is your multi-modal model an all-around player?},
  author={Liu, Yuan and Duan, Haodong and Zhang, Yuanhan and Li, Bo and Zhang, Songyang and Zhao, Wangbo and Yuan, Yike and Wang, Jiaqi and He, Conghui and Liu, Ziwei and others},
  booktitle={European conference on computer vision},
  pages={216--233},
  year={2024},
  organization={Springer}
}

@article{guo2026topoagent,
  title={TopoAgent: A Structure-Aware Perception-to-Reasoning Framework for Diagram-to-Graph Topology Extraction with Large Vision-Language Models},
  author={Guo, Bangwei and Zhao, Xujiang and Liu, Yanchi and Cheng, Wei and Chen, Shengyu and Li, Dongyue and Morimoto, Masaharu and Kuroda, Takayuki and Metaxas, Dimitris and Chen, Haifeng},
  journal={arXiv preprint arXiv:2608.28701},
  year={2026}
}

@article{li2024visiongraph,
  title={Visiongraph: Leveraging large multimodal models for graph theory problems in visual context},
  author={Li, Yunxin and Hu, Baotian and Shi, Haoyuan and Wang, Wei and Wang, Longyue and Zhang, Min},
  journal={arXiv preprint arXiv:2405.04950},
  year={2024}
}

@inproceedings{zheng2025lvlms,
  title={Why lvlms are more prone to hallucinations in longer responses: The role of context},
  author={Zheng, Ge and Qian, Jiaye and Tang, Jiajin and Yang, Sibei},
  booktitle={2025 IEEE/CVF International Conference on Computer Vision (ICCV)},
  pages={4101--4113},
  year={2025},
  organization={IEEE}
}

@article{jiang2026kvsmooth,
  title={Kvsmooth: Mitigating hallucination in multi-modal large language models through key-value smoothing},
  author={Jiang, Siyu and Chen, Feiyang and Zhang, Xiaojin and He, Kun},
  journal={arXiv preprint arXiv:2602.04268},
  year={2026}
}

@inproceedings{yue2024mmmu,
  title={Mmmu: A massive multi-discipline multimodal understanding and reasoning benchmark for expert agi},
  author={Yue, Xiang and Ni, Yuansheng and Zhang, Kai and Zheng, Tianyu and Liu, Ruoqi and Zhang, Ge and Stevens, Samuel and Jiang, Dongfu and Ren, Weiming and Sun, Yuxuan and others},
  booktitle={Proceedings of the IEEE/CVF conference on computer vision and pattern recognition},
  pages={9556--9567},
  year={2024}
}

@article{tang2026chartmuseum,
  title={Chartmuseum: Testing visual reasoning capabilities of large vision-language models},
  author={Tang, Liyan and Kim, Grace and Zhao, Xinyu and Lake, Thom and Ding, Wenxuan and Yin, Fangcong and Singhal, Prasann and Wadhwa, Manya and Liu, Zeyu and Sprague, Zayne and others},
  journal={Advances in Neural Information Processing Systems},
  volume={38},
  year={2026}
}

@article{ai2025graph,
  title={Graph-to-vision: Multi-graph understanding and reasoning using vision-language models},
  author={Ai, Qihang and Li, Ruizhou and Wang, Menghui and Jiang, Haiyun},
  journal={arXiv preprint arXiv:2503.21435},
  year={2025}
}

@article{ouyang2026benchmarking,
  title={Benchmarking and Evaluating VLMs for Software Architecture Diagram Understanding},
  author={Ouyang, Shuyin and Zhang, Jie M and Gong, Jingzhi and Jahangirova, Gunel and Mousavi, Mohammad Reza and Johns, Jack and Lee, Beum Seuk and Ziolkowski, Adam and Virginas, Botond and Noppen, Joost},
  journal={arXiv preprint arXiv:2604.04009},
  year={2026}
}

@article{kumar2026enginuity,
  title={Enginuity: A Dataset and Benchmark for Vision-Language Understanding of Engineering Diagrams},
  author={Kumar, Abhishek and Motiyani, Isha and Kasturi, Tilak and Seefried, Ethan and Movva, Prahitha and Ghosal, Tirthankar},
  journal={arXiv preprint arXiv:2606.03410},
  year={2026}
}

@article{su2026vcg,
  title={VCG-Bench: Towards A Unified Visual-Centric Benchmark for Structured Generation and Editing},
  author={Su, Xiaoyan and Dong, Peijie and Tang, Zhenheng and Tang, Song and Zhai, Yuyao and Lin, Kaitao and Chen, Liang and Yuhang, Gai and Luo, Yuyu and Wang, Qiang and others},
  journal={arXiv preprint arXiv:2605.15677},
  year={2026}
}

@inproceedings{li2025chain,
  title={Chain-of-region: Visual language models need details for diagram analysis},
  author={Li, Xue and Sun, Yiyou and Cheng, Wei and Zhu, Yinglun and Chen, Haifeng},
  booktitle={International Conference on Learning Representations},
  volume={2025},
  pages={28547--28563},
  year={2025}
}

@article{su2026pixel,
  title={Pixel reasoner: Incentivizing pixel space reasoning via curiosity-driven reinforcement learning},
  author={Su, Alex and Wang, Haozhe and Ren, Weiming and Lin, Fangzhen and Chen, Wenhu},
  journal={Advances in Neural Information Processing Systems},
  volume={38},
  pages={8222--8251},
  year={2026}
}

@inproceedings{liu2026small,
  title={Small drafts, big verdict: Information-intensive visual reasoning via speculation},
  author={Liu, Yuhan and Qin, Lianhui and Wan, Shenji},
  booktitle={International Conference on Learning Representations},
  volume={2026},
  pages={5385--5412},
  year={2026}
}

@inproceedings{ye2025beyond,
  title={Beyond end-to-end vlms: Leveraging intermediate text representations for superior flowchart understanding},
  author={Ye, Junyi and Dash, Ankan and Yin, Wenpeng and Wang, Guiling},
  booktitle={Proceedings of the 2025 Conference of the Nations of the Americas Chapter of the Association for Computational Linguistics: Human Language Technologies (Volume 1: Long Papers)},
  pages={3534--3548},
  year={2025}
}

@inproceedings{zeng2026davinci,
  title={DaVinci: Reinforcing Visual-Structural Syntax in MLLMs for Generalized Scientific Diagram Parsing},
  author={Zeng, Xingchen and Su, Zhewei and Zhang, Hengming and Jiang, Juyong and Xia, Jiazhi and Zeng, Wei},
  booktitle={International Conference on Learning Representations},
  volume={2026},
  pages={88664--88699},
  year={2026}
}

@article{bai2025qwen3,
  title={Qwen3-vl technical report},
  author={Bai, Shuai and Cai, Yuxuan and Chen, Ruizhe and Chen, Keqin and Chen, Xionghui and Cheng, Zesen and Deng, Lianghao and Ding, Wei and Gao, Chang and Ge, Chunjiang and others},
  journal={arXiv preprint arXiv:2511.21631},
  year={2025}
}

@article{hong2025glm,
  title={Glm-4.5 v and glm-4.1 v-thinking: Towards versatile multimodal reasoning with scalable reinforcement learning},
  author={Hong, Wenyi and Yu, Wenmeng and Gu, Xiaotao and Wang, Guo and Gan, Guobing and Tang, Haomiao and Cheng, Jiale and Qi, Ji and Ji, Junhui and Pan, Lihang and others},
  journal={arXiv preprint arXiv:2507.01006},
  year={2025}
}

@article{yue2025mimo,
  title={Mimo-vl technical report},
  author={Yue, Zihao and Lin, Zhenru and Song, Yifan and Wang, Weikun and Ren, Shuhuai and Gu, Shuhao and Li, Shicheng and Li, Peidian and Zhao, Liang and Li, Lei and others},
  journal={arXiv preprint arXiv:2506.03569},
  year={2025}
}

@article{yang2025kwai,
  title={Kwai keye-vl 1.5 technical report},
  author={Yang, Biao and Wen, Bin and Ding, Boyang and Liu, Changyi and Chu, Chenglong and Song, Chengru and Rao, Chongling and Yi, Chuan and Li, Da and Zang, Dunju and others},
  journal={arXiv preprint arXiv:2509.01563},
  year={2025}
}

@article{wang2025internvl3,
  title={Internvl3. 5: Advancing open-source multimodal models in versatility, reasoning, and efficiency},
  author={Wang, Weiyun and Gao, Zhangwei and Gu, Lixin and Pu, Hengjun and Cui, Long and Wei, Xingguang and Liu, Zhaoyang and Jing, Linglin and Ye, Shenglong and Shao, Jie and others},
  journal={arXiv preprint arXiv:2508.18265},
  year={2025}
}

@article{lu2025ovis2,
  title={Ovis2. 5 technical report},
  author={Lu, Shiyin and Li, Yang and Xia, Yu and Hu, Yuwei and Zhao, Shanshan and Ma, Yanqing and Wei, Zhichao and Li, Yinglun and Duan, Lunhao and Zhao, Jianshan and others},
  journal={arXiv preprint arXiv:2508.11737},
  year={2025}
}

@article{hurst2024gpt,
  title={Gpt-4o system card},
  author={Hurst, Aaron and Lerer, Adam and Goucher, Adam P and Perelman, Adam and Ramesh, Aditya and Clark, Aidan and Ostrow, AJ and Welihinda, Akila and Hayes, Alan and Radford, Alec and others},
  journal={arXiv preprint arXiv:2410.21276},
  year={2024}
}

@article{singh2025openai,
  title={Openai gpt-5 system card},
  author={Singh, Aaditya and Fry, Adam and Perelman, Adam and Tart, Adam and Ganesh, Adi and El-Kishky, Ahmed and McLaughlin, Aidan and Low, Aiden and Ostrow, AJ and Ananthram, Akhila and others},
  journal={arXiv preprint arXiv:2601.03267},
  year={2025}
}

@misc{anthropic2026opus5,
  author       = {{Anthropic}},
  title        = {Claude Opus 5},
  year         = {2026},
  howpublished = {\url{https://platform.claude.com/docs/en/models/opus-5/whats-new-opus-5}},
  note         = {Claude Platform Documentation, accessed September 2026}
}
\bibliographystyle{iclr2027_conference}

\appendix
\section{Related Work}

\paragraph{Benchmarks for structural diagram understanding.}
Existing multimodal and diagram-understanding benchmarks largely formulate
structured visual understanding as question answering or reasoning over
selected visual content. Representative examples include AI2D
~\citep{kembhavi2016diagram}, ChartQA~\citep{masry2022chartqa},
MMBench~\citep{liu2024mmbench}, MMMU~\citep{yue2024mmmu}, and
ChartMuseum~\citep{tang2026chartmuseum}. More recent work moves toward
graph-centric visual reasoning: VisionGraph~\citep{li2024visiongraph},
VGCURE~\citep{zhu2025benchmarking}, and Graph-to-Vision
~\citep{ai2025graph} evaluate structural and relational reasoning over
visually rendered graphs, while domain-specific benchmarks study structured
understanding and extraction in software architecture, engineering, and other
specialized diagrams
~\citep{ouyang2026benchmarking,kumar2026enginuity,su2026vcg}. Synthetic data has also been used to scale structured diagram supervision.
SSVC~\citep{scholch2022towards} provides synthetic structured-visual-content
images with bounding-box and graph annotations, while FlowLearn
~\citep{pan2024flowlearn} and FlowGen~\citep{shi2026flowgen} use simulated or
controllably generated flowcharts for structured understanding and evaluation.
Most closely related,
\mbox{TopoBench-180~\citep{guo2026topoagent}} provides a
human-verified real-world evaluation set with canonical graph annotations for
diagram-to-graph topology extraction. In contrast, Knossos provides
large-scale fully supervised training and controlled evaluation across six
domains, with complete topology annotations and systematic variation in
topology, layout, connector routing, and visual appearance.

\paragraph{Structure-aware diagram understanding and topology extraction.}
Recent work on diagram understanding has increasingly moved from localized
visual evidence toward explicit structural representations. Chain-of-Region
decomposes scientific diagrams into semantically meaningful regions for
fine-grained reasoning~\citep{li2025chain}, while related visual-reasoning
methods such as Pixel Reasoner~\citep{su2026pixel} and Speculative
Verdict~\citep{liu2026small} improve detailed perception through pixel-space
reasoning and speculative visual inference. Moving beyond localized evidence,
TextFlow~\citep{ye2025beyond} converts flowcharts into textual graph representations before
downstream reasoning, Diagram2Structure~\citep{hu2026diagram2structure} reconstructs raster diagrams into
standardized structured representations, and DaVinci~\citep{zeng2026davinci} learns diagram structure
from visual primitives and their relations. More directly related, TopoAgent~\citep{guo2026topoagent}
targets diagram-to-graph topology extraction through grounded structural
prediction. Ariadne builds on this direction through supervised complete-topology training
and source-group decomposition, with local predictions deterministically
assembled into the final graph.

\section{Detailed Benchmark Construction}
\label{app:benchmark_details}

This appendix describes the benchmark construction pipeline in sufficient
detail for reproduction. We first introduce the domain-specific node
vocabularies and graph grammars used to generate valid symbolic topologies.
We then describe visual-asset generation and sampling, layout and connector
routing, annotation derivation, quality control, and the final balancing and
release procedure. We additionally specify the boundary between the released
synthetic benchmark and the separately curated real-diagram evaluation set.

\subsection{Canonical vocabularies and node instances}
\label{app:node_vocabularies}

Each domain is defined by a version-controlled JSON vocabulary of canonical
node entries. An entry separates a stable key and base display name from the
attributes used by generation, including a broader semantic category and, when
applicable, a structural role, preferred shape, render mode, or symbol. For
example, one Circuit entry is
\texttt{\{label: ground, display: Ground, category: reference, role: ground,
symbol: ground\}}. The graph grammar uses its category and role to place the
node in a valid topology, while the renderer uses its display name and symbol
to construct its appearance. Keeping these vocabularies with the released
generator fixes the available node semantics for a dataset version and makes
later vocabulary changes explicit.

Instantiating a vocabulary entry creates a diagram-specific node. To avoid
repeating a small closed set of labels verbatim, the renderer may append a
domain-specific suffix sampled from letters, numbers, or location terms, as in
\texttt{Server 03} or \texttt{Server West}. If the same base label occurs more
than once in a diagram, distinct suffixes are assigned to disambiguate the
instances, as in \texttt{Ground X} and \texttt{Ground 2}. Capitalization may
also vary independently. Every instance stores the exact label drawn in the
image, a unique diagram-local ID, and a bounding box; these surface variations
do not change its underlying vocabulary entry or structural role.

The frozen release contains 310 domain-local canonical labels across the six
domains. Table~\ref{tab:node_vocabulary_stats} summarizes their coverage and the
resulting node counts.

\begin{table}[h]
\caption{Canonical vocabulary and node-occurrence statistics in the frozen
release. Domain identifies the six diagram domains; Canonical
labels is the number of domain-local vocabulary entries represented in that
domain; Node instances is the total number of node occurrences across
its 3,200 diagrams; Avg./image is the mean number of nodes per diagram;
and Min--max is the observed per-diagram node-count range. The total row
aggregates all 19,200 diagrams, counting a shared label separately in each
domain.}
\label{tab:node_vocabulary_stats}
\centering
\small
\begin{tabular}{lrrrr}
\toprule
Domain & Canonical labels & Node instances & Avg./image & Min--max \\
\midrule
Food Web & 31 & 38,585 & 12.06 & 6--17 \\
Network & 40 & 44,315 & 13.85 & 8--17 \\
Workflow & 60 & 47,746 & 14.92 & 8--20 \\
Natural Process & 59 & 40,527 & 12.66 & 7--18 \\
Circuit & 60 & 45,142 & 14.11 & 7--23 \\
Map Route & 60 & 28,864 & 9.02 & 6--15 \\
\midrule
Total & 310 & 245,179 & 12.77 & 6--23 \\
\bottomrule
\end{tabular}
\end{table}

\paragraph{Food Web (31 names; three trophic tiers).}
\emph{Producer (8):} Grass, Berries, Algae, Leaves, Acorn, Flowering Plant,
Plankton, Seaweed.
\emph{Intermediate consumer (12):} Rabbit, Mouse, Squirrel, Deer, Caribou,
Moose, Caterpillar, Grasshopper, Dragonfly, Butterfly, Frog, Fish.
\emph{Higher consumer (11):} Bobcat, Fox, Hawk, Owl, Gray Wolf, Black Bear,
Snake, Eagle, Shark, Seal, Killer Whale.
The FoodWeb appearance pool also contains assets for Mushroom, Coral, Shrimp,
Clam, Snail, Spruce Tree, Evergreen Tree, Oak Tree, and Pine Tree. These nine
candidate labels are not sampled by the frozen graph generator and therefore
are not counted in the 31-name benchmark vocabulary.

\paragraph{Network (40 names; 15 categories).}
\emph{Compute (1):} Server.
\emph{Endpoint (7):} Client, Workstation, Laptop, Desktop, Phone, Tablet,
Terminal.
\emph{Infrastructure (1):} Data Center.
\emph{IoT (3):} Sensor, Controller, IoT Device.
\emph{Network device (4):} Repeater, Bridge, Modem, Relay.
\emph{Network group (2):} Subnet, Cluster.
\emph{Network node (2):} Edge Node, Core Node.
\emph{Peripheral (2):} Printer, Camera.
\emph{Routing (2):} Router, Gateway.
\emph{Security (4):} Firewall, VPN, IDS, IPS.
\emph{Service (2):} DNS Server, Cloud.
\emph{Storage (3):} Database, Storage, NAS.
\emph{Switching (2):} Switch, Hub.
\emph{Traffic (2):} Load Balancer, Proxy.
\emph{Wireless (3):} Access Point, Base Station, Satellite.

\paragraph{Workflow (60 names; 10 categories).}
\emph{Terminal (2):} Start, End.
\emph{Intake (3):} Receive Request, Submit Form, Capture Input.
\emph{Validation (7):} Validate Input, Check Completeness, Verify Identity,
Check Eligibility, Check Inventory, Quality Check, Run Test.
\emph{Data pipeline (8):} Extract Data, Transform Data, Load Data, Query
Database, Update Database, Sync Records, Generate Report, Export File.
\emph{Automation (8):} Run Model, Score Result, Classify Item, Route Task,
Enqueue Job, Process Job, Retry Step, Handle Error.
\emph{Approval (7):} Manual Review, Review Result, Approve Request, Reject
Request, Escalate Case, Assign Owner, Sign Off.
\emph{Communication (6):} Send Notification, Send Email, Create Ticket, Update
Status, Notify Customer, Schedule Meeting.
\emph{Operations (6):} Pack Item, Ship Order, Receive Shipment, Inspect Item,
Repair Item, Close Case.
\emph{Storage (4):} Archive Record, Store Document, Retrieve Record, Backup
Data.
\emph{Control flow (9):} Merge Branches, Split Path, Wait for Event, Timer
Expires, Parallel Tasks, Join Tasks, Decision Point, Condition Met?, Fallback
Path.

\paragraph{Natural Process (59 realized names; nine categories).}
\emph{Energy (2):} Sunlight, Heat.
\emph{Climate (2):} Wind, Cloud.
\emph{Water (8):} Rain, Snow, River, Stream, Lake, Ocean, Groundwater, Glacier.
\emph{Earth (6):} Soil, Sand, Clay, Rock, Mineral, Sediment.
\emph{Organic (4):} Leaf Litter, Organic Matter, Dead Wood, Compost.
\emph{Organism (14):} Plant, Grass, Tree, Algae, Moss, Fungi, Bacteria,
Microbe, Pollinator, Insect, Worm, Fish, Bird, Small Animal.
\emph{Role (1):} Decomposer.
\emph{Chemical (8):} Carbon Dioxide, Oxygen, Nitrogen, Phosphate, Nitrate,
Ammonium, Nutrient, Toxin.
\emph{Process (14):} Photosynthesis, Respiration, Decomposition, Evaporation,
Condensation, Runoff, Infiltration, Sedimentation, Erosion, Absorption,
Fixation, Mineralization, Uptake, Release.
The candidate configuration also contains the human--nature label Pollution,
but it is not selected by the frozen layer pools and does not occur in the
19,200 released annotations.

\paragraph{Circuit (60 names; 11 categories).}
\emph{Power source (5):} Battery, Power Supply, USB Power, Solar Cell, AC
Adapter.
\emph{Power conditioning (3):} Voltage Regulator, Buck Converter, Boost
Converter.
\emph{Protection (1):} Fuse.
\emph{Passive (7):} Resistor, Variable Resistor, Capacitor, Polarized
Capacitor, Inductor, Transformer, Crystal Oscillator.
\emph{Passive sensor (2):} Thermistor, Photoresistor.
\emph{Active (10):} Diode, LED, Zener Diode, Transistor, MOSFET, Op Amp,
Comparator, Logic Gate, Integrated Circuit, Driver Chip.
\emph{Control (6):} Switch, Push Button, Toggle Switch, Relay, Microcontroller,
Arduino Board.
\emph{Connector (6):} DC Jack, Jumper, GPIO Pin, Terminal Block, Header
Connector, Test Point.
\emph{Sensor (10):} Temperature Sensor, Light Sensor, Motion Sensor, Pressure
Sensor, Humidity Sensor, Current Sensor, Voltage Sensor, Hall Sensor,
Microphone, Photodiode.
\emph{Load (9):} Motor, Servo Motor, Speaker, Buzzer, Lamp, Heater, Display,
Fan, Pump.
\emph{Reference (1):} Ground.

\paragraph{Map Route (60 names; six categories).}
\emph{Transit (10):} Station, Platform, Terminal, Transfer Hub, Depot, Bus
Stop, Train Stop, Tram Stop, Ferry Pier, Taxi Stand.
\emph{Indoor (10):} Lobby, Office, Laboratory, Storage Room, Elevator,
Stairwell, Hallway, Conference Room, Reception, Server Room.
\emph{Campus (10):} Main Gate, Library, Dormitory, Cafeteria, Parking Lot,
Auditorium, Gymnasium, Courtyard, Admin Building, Lecture Hall.
\emph{Warehouse (10):} Loading Dock, Packing Area, Shelf Zone, Conveyor,
Sorting Area, Cold Storage, Forklift Bay, Inventory Desk, Dispatch Gate,
Return Station.
\emph{Landmark (10):} Bridge, Park, Tower, Plaza, Checkpoint, Fountain, Museum,
Monument, Tunnel Entrance, Overlook.
\emph{Utility (10):} Junction, Entrance, Exit, Security Post, Control Room,
Service Gate, Emergency Exit, Information Desk, Rest Area, Access Point.

\subsection{Domain graph grammars}
\label{app:domain_grammars}

Each domain grammar begins from semantic node roles and a collection of valid
structural motifs rather than sampling unconstrained random graphs. Difficulty
then changes graph scale, density, and geometric ambiguity in ways appropriate
to that domain. The statistics below are measured from the frozen release.

\begin{figure}[t]
    \centering
    \includegraphics[width=\linewidth]{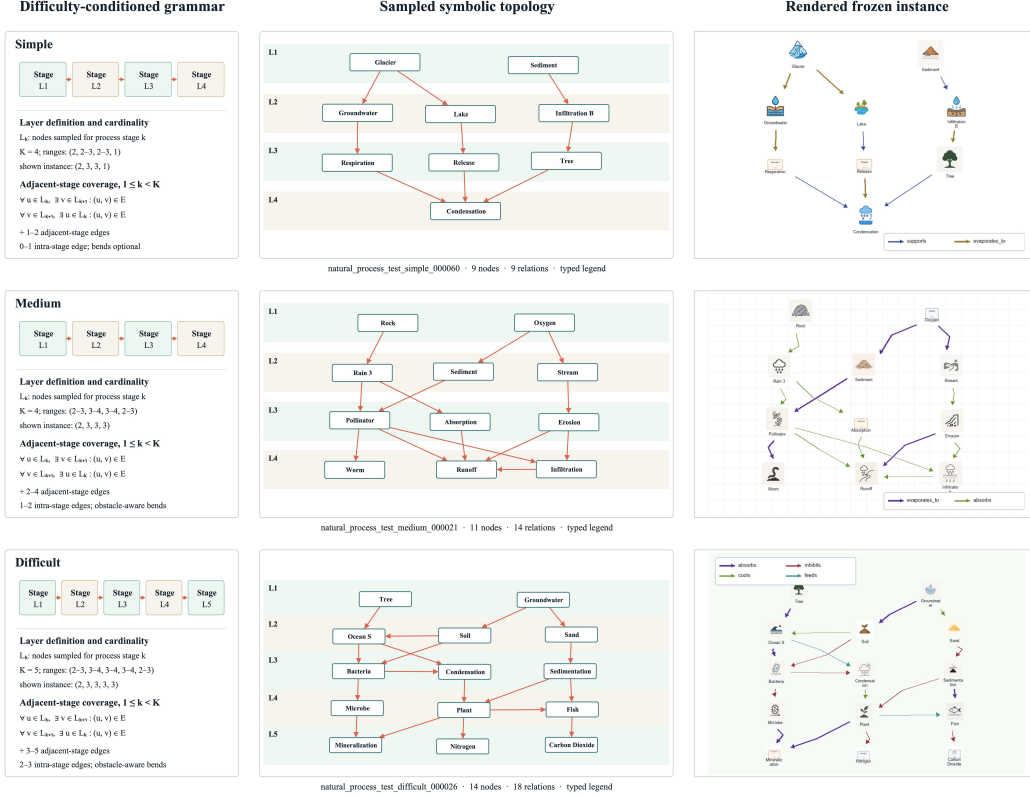}
    \caption{Difficulty-conditioned Natural Process grammar and three frozen
    instantiations. Left: the executable constraints used at each difficulty.
    Here $L_k$ is the node set at process stage $k$, $K$ is the number of
    stages, and the two coverage constraints require every node in adjacent
    stages to participate in at least one forward relation. The listed ranges
    are grammar-level cardinality ranges, while ``shown instance'' gives the
    realized cardinalities in that row. Middle: the exact symbolic nodes and
    edge multiset of the selected example, with geometry regularized only for
    legibility. Right: the corresponding unmodified rendered image and its
    relation-type legend. From top to bottom, the examples contain
    9/9, 11/14, and 14/18 nodes/relations. All three are drawn directly from
    the frozen test set; no topology, annotation, or rendered connector was
    edited for the figure.}
    \label{fig:natural_process_grammar_scaling}
\end{figure}

Across domains, generation follows the same high-level procedure: sample
difficulty-conditioned grammar parameters and role slots, instantiate a
symbolic graph satisfying the resulting structural constraints, and only then
assign layout, connector routing, and visual assets.
Figure~\ref{fig:natural_process_grammar_scaling} illustrates this procedure
with the layered Natural Process grammar. Let $L_k$ denote the nodes sampled
for process stage $k$. For every adjacent pair
$L_k,L_{k+1}$, the generator gives each node in $L_k$ at least one outgoing
edge into $L_{k+1}$ and each node in $L_{k+1}$ at least one incoming edge from
$L_k$, before adding difficulty-dependent adjacent-stage and intra-stage
relations. The example demonstrates how difficulty modifies the grammar---the
number and cardinality of stages, relation density, and routing complexity---
rather than merely enlarging an otherwise fixed graph.

\paragraph{Food Web.}
The grammar samples producer, intermediate-consumer, and higher-consumer tiers
and arranges them in one of four orientations. In the frozen release, simple
images contain 6--8 nodes and 5--9 edges, medium images contain 9--13 nodes and
7--15 edges, and difficult images contain 16--17 nodes and 14--22 edges. Most
relations follow plausible trophic direction between adjacent tiers, while
denser examples add long-range links and directed, bidirectional, or unarrowed
connector variants. Medium and difficult cases may also contain close visual
alternatives and connector crossings that do not denote additional relations.
Node, label, and arrow sizes are reduced together in dense layouts so that
difficulty is not produced by simply adding edges to an unchanged canvas.

\paragraph{Network.}
Network diagrams combine hub--spoke, layered infrastructure, tiered, and
cluster-bridge structures, with device roles constraining plausible
connections. Simple examples contain 8--10 nodes and 8--12 edges, medium
examples 12--15 nodes and 13--21 edges, and difficult examples 14--17 nodes and
23--31 edges. Increasing difficulty introduces multiple hubs or clusters,
shorter distances between candidate endpoints, and more links sharing the same
devices or bridge regions. Connectors primarily express unarrowed structural
connectivity, with routing adjusted in denser cases to keep connectivity
visually traceable.

\paragraph{Workflow.}
Workflow templates encode three control-flow families used in the frozen
release: staged grids, ladder-feedback workflows, and swimlanes. Each template
defines role slots and a backbone that preserves the functions of terminals,
processes, decisions, splits, and merges.
Simple, medium, and difficult examples contain 8--9, 13--14, and 20 nodes,
respectively, with 10--11, 18, and 29--32 directed edges. Harder cases increase
the number of stages or parallel lanes and may introduce skip or feedback
relations. Role-specific shapes are rendered locally, while typed connector
styles are assigned after topology creation and communicated through a matching
legend.

\paragraph{Natural Process.}
Natural Process diagrams organize sources, carriers, reservoirs, flows,
organisms, and materials into layered transformation or transport structures.
Simple and medium cases use four layers with 7--9 and 10--14 nodes, while
difficult cases use five layers with 13--18 nodes. Their corresponding edge
ranges are 6--13, 10--23, and 15--33. The grammar first establishes progression
between adjacent layers, ensures that nodes participate in the process, and
then adds difficulty-dependent cross-layer or reversible within-layer links.
An image may use one or several semantic relation types and may contain
multiple typed relations between the same endpoint pair. Parallel connectors
are visually separated and distinguished through the legend.

\paragraph{Circuit.}
Circuit diagrams use layered functional structures that progress from power,
sensing, or passive components through control and active components to loads
and grounds. Simple and medium diagrams use four layers, whereas difficult
diagrams use five; the frozen ranges are 7--10, 10--14, and 16--23 nodes and
6--12, 9--22, and 18--35 edges. Endpoint roles determine whether a relation is
rendered as wire, power, ground, or signal: links to ground use the ground
relation, power-source links favor power, and control, sensor, or active
components favor signal or wire. Harder diagrams add denser inter-layer and
parallel typed connections. Shared endpoints and parallel links are visually
separated and identified in a legend.

\paragraph{Map Route.}
Map Route diagrams cover city and campus grids, indoor corridors, transit
networks, subway-like lines, floorplans, warehouse routes, cluster bridges, and
free spatial maps. Simple examples contain 6--8 nodes and 4--7 edges, medium
examples contain 8--12 nodes and 6--16 edges, and difficult examples contain
8--15 nodes and 8--25 edges. Difficulty additionally increases intersections,
shared corridors, and alternative paths between nearby destinations. Routes
may be direct, softly bent, or orthogonal and may be directed or unarrowed.
Overlapping routes and parallel links are visually separated, while route types
and direction conventions are indicated by the legend.

\subsection{Visual asset pipeline}
\label{app:asset_pipeline}

Visual appearance is sampled only after the symbolic graph has been fixed.
The graph generator determines node identities and roles, typed relations, and
layout constraints; the asset pipeline supplies alternative appearances for
the resulting nodes. Consequently, changing an asset cannot change the graph
or any evaluation target.

\begin{figure}[t]
    \centering
    \includegraphics[width=\linewidth]{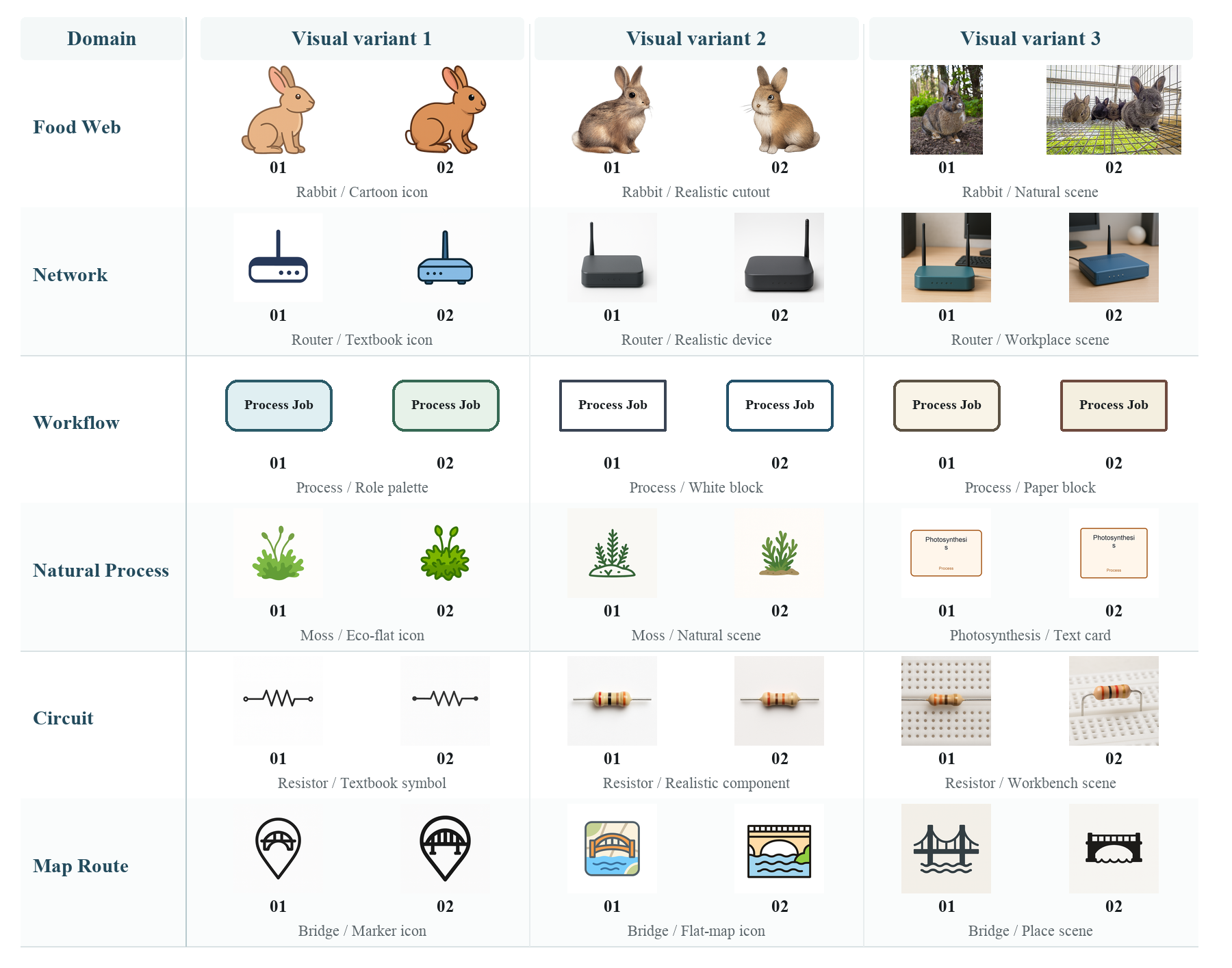}
    \caption{Representative visual-appearance pools in Knossos. Each row is a
    domain, each column is an appearance family, and each cell shows the two
    candidate variants available for one representative vocabulary entry.
    Workflow appearances are rendered programmatically rather than generated.
    The figure shows the candidate pool; the frozen renderer uses the
    domain-specific subset and sampling policy described below.}
    \label{fig:appendix_asset_pools}
\end{figure}

\paragraph{Asset construction.}
For each image-eligible vocabulary entry, the pipeline constructs
label--style--variant generation jobs from its canonical label, semantic role,
and requested visual style. Prompts request a single centered subject in a
square composition and explicitly exclude text, arrows, diagram lines, brands,
borders, and watermarks. Two independent candidates are generated for each
label--style pair and screened for recognizability, unwanted text, and visual
quality. Assets are generated with the OpenAI Images API using
\texttt{gpt-image-1-mini} at $1024\!\times\!1024$ resolution, medium quality,
and PNG output. Visible node labels are rendered separately from the symbolic
annotations using Pillow rather than by the image model, ensuring exact
agreement between displayed text and stored ground truth. The
$1024\!\times\!1024$ resolution above refers only to source assets before they
are fitted into diagram nodes, not to the resolution of the rendered diagrams.

\paragraph{Domain-specific visual pools.}
Food Web uses three appearance families over 40 candidate labels, of which 31
occur in the frozen graph vocabulary. Network uses textbook-icon and
realistic-device assets. Natural Process combines two generated icon families
for image-eligible concepts with locally rendered text cards for abstract
process nodes. Circuit and Map Route each use three visual families over their
60-category vocabularies. Workflow does not use generated image assets; its
nodes are rendered programmatically using role-specific shapes. Representative
examples are shown in Figure~\ref{fig:appendix_asset_pools}. The resulting
candidate pools contain 240 Food Web, 240 Network, 212 Natural Process, 360
Circuit, and 360 Map Route assets.

\paragraph{Sampling in the frozen renderer.}
Asset sampling is domain-specific and seeded for reproducibility. Food Web
samples independently from label-matched visual candidates for each node,
whereas Network selects a consistent visual style at the image level. Natural
Process, Circuit, and Map Route alternate between fixed-style and mixed-style
rendering to increase visual diversity. Map Route additionally includes
locally rendered text-only nodes. The selected asset identity is stored with
each node annotation to support exact replay.

\subsection{Geometry and annotation derivation}
\label{app:geometry_annotations}

Let $B_i$ denote the finalized bounding box of node $v_i$, and let
$e=(s,t)$ denote a directed edge, optionally associated with a relation type
$r$. The renderer obtains source and target anchors by intersecting the
center-to-center ray with the boundaries of $B_s$ and $B_t$. It then constructs
a logical connector polyline $P_e$ using the routing family associated with the
sampled domain and template: direct, softly bent, or orthogonal. When
obstacle-aware routing is used, candidate detours are selected according to
their interference with non-endpoint nodes and path length.

Terminal clearance and arrow placement may slightly shorten the path actually
drawn. We therefore store both the logical polyline $P_e$ and the rendered
polyline $\widetilde P_e$ (\texttt{visible\_path}). Arrowheads are derived from
the terminal direction of $\widetilde P_e$, and reciprocal directions use the
same path in reverse. Any image-resizing transform is applied consistently to
node boxes, centers, path vertices, and arrowheads, preserving
image--annotation alignment.

\begin{figure}[t]
    \centering
    \includegraphics[width=0.58\linewidth]{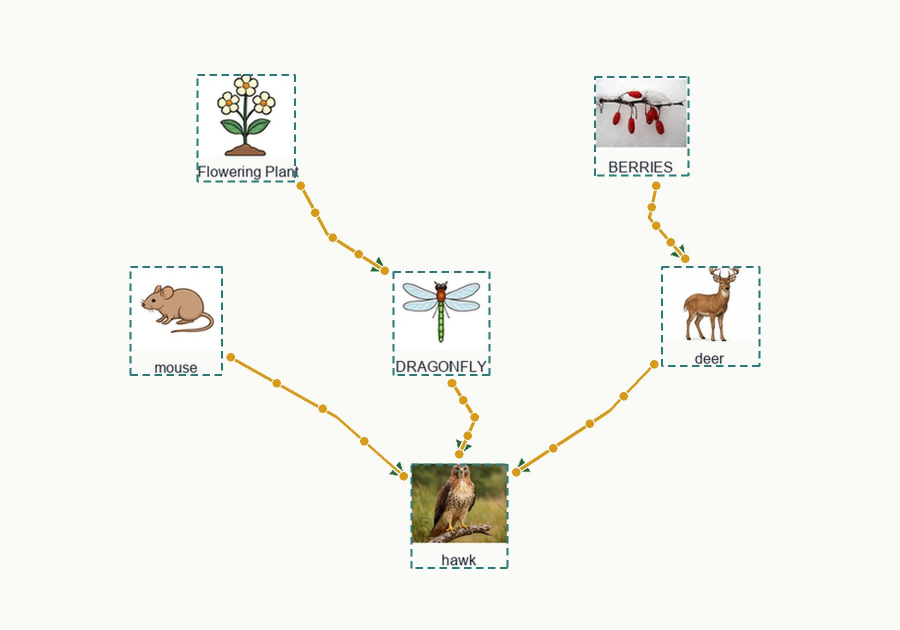}
    \caption{Geometry annotations on a released Food Web sample. Teal dashed
    boxes denote finalized node bounding boxes. Gold polylines denote rendered
    connector paths (\texttt{visible\_path}), and the five circular markers on
    each path are trace points sampled at equal arc-length intervals in
    source-to-target order. All coordinates are stored in pixels on the final
    image canvas.}
    \label{fig:geometry_annotation_example}
\end{figure}

A five-point trace is derived deterministically from the rendered path. Let
$\widetilde p_e:[0,1]\rightarrow\mathbb{R}^2$ denote its normalized
arc-length parameterization in the final image coordinate system. The parameter
$t\in[0,1]$ specifies relative traversal distance along the connector, while
$\widetilde p_e(t)$ remains an $(x,y)$ location on the resized image canvas.
The geometric target is
\begin{equation}
    \tau_e = \bigl(\widetilde p_e(j/4)\bigr)_{j=0}^{4}.
    \label{eq:trace5_derivation}
\end{equation}
The sampled coordinates are rounded to the nearest integer pixel. For the
reciprocal direction, the five trace points are stored in reverse order. Node
records, edges, relation types when present, connector geometry, and training
targets are all serialized from the same finalized objects; none is recovered
from rendered pixels or entered manually. Figure~\ref{fig:geometry_annotation_example} visualizes these annotations on a
released Food Web sample. Every node is overlaid with its finalized bounding
box, and every connector with its five sampled trace points.

\subsection{Quality control and release construction}
\label{app:qc_release}

Before large-scale generation, human researchers inspect preview batches from
each domain to verify that the generated diagrams are visually coherent and
semantically plausible. The review focuses on overall image quality, node and
connector readability, layout and routing plausibility, consistency with the
intended domain structure, and whether increasing difficulty produces
meaningfully more challenging yet still interpretable diagrams. Generation
rules and rendering configurations are refined when systematic issues are
identified.

After the generation templates and rendering rules are validated through human
inspection, the pipeline performs large-scale generation, producing 24,000
candidate diagrams across the six domains, with 4,000 candidates generated for
each domain. Visual screening is then applied directly to the rendered images
using GPT-4o with temperature zero. The binary screening criteria cover overall
clarity, text readability, node and connector visibility, arrow and legend
clarity, and excessive overlap. Dense layouts, crossings, and mild local
clutter are allowed as long as the diagram remains visually interpretable,
preventing difficult examples from being systematically filtered out. Visual screening continued until all release quotas were filled. In total,
GPT-4o evaluated 19,223 candidate diagrams, of which 19,200 passed and 23
failed. 

For release construction, candidates are deterministically shuffled within each
domain and difficulty level and assigned to training and test pools before
screening. The selected release contains 3,000 training and 200 test images per
domain, yielding 18,000 training images and 1,200 held-out test images in total,
with matching JSON annotations and manifest entries. Released diagrams use one
of two fixed resolutions: $900\times630$ pixels for landscape layouts and
$630\times900$ pixels for portrait layouts. Of the released images, 14,318
(74.6\%) use the landscape resolution and 4,882 (25.4\%) use the portrait
resolution. The canvas dimensions are stored in every annotation, and all node
boxes, connector paths, arrowheads, and trace points are defined in this final
pixel coordinate system. An audit of the complete release found exact agreement
between the recorded canvas dimensions and the actual PNG resolutions.


\subsection{TopoBench-180 evaluation subset}
\label{app:topobench_preprocessing}

We evaluate on a 162-image subset of
TopoBench-180~\citep{guo2026topoagent}, comprising 54 AI2D/Web and 108
Network diagrams. Relative to the original 180-image collection, we
exclude 16 internal Network diagrams and two Network diagrams with
unresolved edge references. These exclusions reflect data availability
and annotation integrity, rather than model performance. We convert
name-keyed annotations to stable node-instance IDs and remove
annotation-only naming suffixes while preserving visible labels and
distinct repeated-label instances. All evaluation images are obtained
from the released files or their provided source links. The resulting
annotations contain 1,786 nodes and 3,063 ordered relations. We use this
subset exclusively for external evaluation and report Node F1 and
Edge F1.

\section{Theoretical Derivation of Source-Group Decomposition}
\label{app:decomposition}

Prior work has observed that longer vision--language generations can be more
susceptible to hallucination and semantic drift
~\citep{zheng2025lvlms,jiang2026kvsmooth}. These observations motivate
examining how Ariadne's source-group decomposition changes both the output
representation and the autoregressive decoding process by replacing one long
graph serialization with multiple shorter generations.

We condition throughout on a fixed node inventory and isolate the
edge-decoding stage; errors introduced by upstream node prediction are outside
the scope of the derivation. Our analysis proceeds from representation to
decoding. We first establish that source grouping preserves the target edge
multiset, quantify the cross-group ordering freedom in a one-shot
representation, and examine its implications for sequence supervision.
We then bound the edge-decoding horizon and, under an explicit
position-sensitive error model, derive bounds on the reduction in cumulative
position-dependent error.

\subsection{Lossless Decomposition and Serialization Ambiguity}

We begin by separating the target topology from the order in which its edges
are serialized. Let
\[
\mathcal{S}(V)=\{S_1,\ldots,S_K\}
\]
be a partition of the node inventory, and let $E_{S_i}$ denote the multiset of
directed edges whose sources belong to $S_i$. Since every directed edge has
a unique source and therefore belongs to exactly one source group,
\begin{equation}
\label{eq:app_lossless_partition}
E
=
\biguplus_{i=1}^{K} E_{S_i}.
\end{equation}
Thus, source grouping is lossless with respect to graph topology: it changes
the decoding representation without restricting the set of edge multisets
that can be represented.

Although the target topology is unchanged, its serialization may still admit
multiple equivalent orderings. Let
\begin{equation}
m_i = |E_{S_i}|,
\qquad
m = |E| = \sum_{i=1}^{K} m_i.
\end{equation}
Fix the within-group edge orders, denoted by $\mathcal O$, and consider a flat
one-shot representation that permits arbitrary cross-group interleavings.
All other serialization choices are held fixed. An interleaving is determined
by assigning $m_i$ of the $m$ output positions to each group; the fixed
within-group orders then determine which edge occupies each assigned position.
The number of equivalent interleavings is therefore
\begin{equation}
\label{eq:app_serialization_ambiguity}
N_{\mathrm{interleave}}
=
\binom{m}{m_1,\ldots,m_K}
=
\frac{m!}{\prod_{i=1}^{K} m_i!}.
\end{equation}

For this one-shot representation, we quantify cross-group ordering freedom
by its log-cardinality,
\begin{equation}
\label{eq:app_serialization_entropy}
A_{\mathrm{one}}
=
\log N_{\mathrm{interleave}}
=
\log m!
-
\sum_{i=1}^{K}\log m_i!,
\end{equation}
where $\log$ denotes the natural logarithm. This quantity measures the size
of the equivalence class, not the entropy of a particular model distribution.
Ariadne instead decodes each $E_{S_i}$ in a separate call and combines the
results by multiset union. Since edges from different groups are not
interleaved within a shared output sequence, its corresponding cross-group
ambiguity is
\begin{equation}
A_{\mathrm{group}}=0,
\end{equation}
giving
\begin{equation}
\label{eq:app_ambiguity_reduction}
\Delta A
=
A_{\mathrm{one}}-A_{\mathrm{group}}
=
\log\frac{m!}{\prod_{i=1}^{K}m_i!}.
\end{equation}
Possible ordering choices within each group remain unchanged.

This comparison establishes a difference in representation, rather than a
difference in prediction error. Its relevance to training arises when
supervision specifies a single reference sequence: multiple orderings may
encode the same topology, but the sequence-level objective evaluates the
probability of the particular reference ordering. We next make this
distinction explicit.

\paragraph{Implication for sequence supervision.}
Fix the input $X=(I,V)$ and retain the within-group orders $\mathcal O$
defined above. Let $\Sigma_{\mathcal O}(E)$ denote the set of
$N_{\mathrm{interleave}}$ equivalent interleavings. For a sequence model
$p_\theta$, define their total probability mass and the conditional
distribution within this class as
\begin{equation}
\begin{aligned}
q
&=
\sum_{y\in\Sigma_{\mathcal O}(E)}p_\theta(y\mid X),\\
\pi_\theta(y\mid E,\mathcal O,X)
&=
\frac{p_\theta(y\mid X)}{q},
\qquad y\in\Sigma_{\mathcal O}(E),
\end{aligned}
\end{equation}
assuming $q>0$. Because the within-group orders are fixed, $q$ is the
probability assigned to this interleaving class, not necessarily to all
serializations of the target graph.

For a reference sequence $y^\star\in\Sigma_{\mathcal O}(E)$ with positive
probability, its negative log-likelihood decomposes exactly as
\begin{equation}
-\log p_\theta(y^\star\mid X)
=
-\log q
-
\log\pi_\theta(y^\star\mid E,\mathcal O,X).
\end{equation}
The first term evaluates the total probability assigned to the equivalent
interleavings. The second is the additional loss incurred by specifying one
reference ordering rather than marginalizing over this class. In the
idealized uniform case,
$\pi_\theta(y^\star\mid E,\mathcal O,X)=1/N_{\mathrm{interleave}}$,
so the additional term equals $\log N_{\mathrm{interleave}}=A_{\mathrm{one}}$.
Otherwise, it depends on the probability assigned to the particular reference.

This decomposition connects the ordering freedom counted above to an
ordering-dependent component of sequence supervision. Source-group decoding
removes the need to choose a cross-group interleaving, while leaving possible
within-group ordering choices intact. It does not, however, guarantee a lower
total training loss or graph-prediction error, because the grouped model may
assign different probabilities to the target outputs. The comparison also
depends on the one-shot format: enforcing a unique cross-group order removes
this particular ordering freedom, whereas canonically ordered training
targets alone do not constrain the model's output support. The analysis so far concerns how topology is represented and supervised.
The following subsections examine a separate consequence of source
grouping---a shorter decoding horizon---and its implications for
position-dependent error.

\subsection{Bound on the Decoding Horizon}

Source grouping also shortens the maximum edge-decoding horizon of each
autoregressive generation. A one-shot representation predicts all $m=|E|$
edges in a single sequence,
\begin{equation}
L_{\mathrm{one}} = m,
\end{equation}
whereas source-group decoding requires at most
\begin{equation}
\label{eq:app_decoding_horizon}
L_{\mathrm{group}}
=
\max_i m_i
\leq m,
\end{equation}
where $m_i=|E_{S_i}|$. The inequality is strict whenever at least two source
groups contain edges. Thus, source grouping preserves the complete edge set
while replacing one global decoding horizon with multiple shorter local ones.
This reduction motivates the position-sensitive error analysis below.

\subsection{Position-Sensitive Decoding Model}

Empirical studies of LVLM generation have reported greater hallucination risk
and semantic drift in longer responses, including increased error toward later
decoding positions
~\citep{zheng2025lvlms,jiang2026kvsmooth}. Motivated by this behavior, we
consider a stylized position-sensitive decoding model.

Let $\epsilon_j$ denote the expected error associated with an edge-level
decision made at position $j$ within one autoregressive generation. For this stylized comparison, we assume that $\epsilon_j$ captures the
position-dependent component of edge-level error and is shared between the
one-shot and grouped decoding processes. We first
assume only monotonicity:
\begin{equation}
\label{eq:app_length_assumption}
\epsilon_{j+1}
\geq
\epsilon_j.
\end{equation}
This assumption does not assert that output position itself causes error.
Rather, it models a regime in which later decisions are no easier on average
than earlier ones, potentially reflecting accumulated uncertainty, semantic
drift, or increasing dependence on previously generated context.

For the comparison below, consider a one-shot serialization in which the
source groups are written as contiguous blocks in the order
$S_1,\ldots,S_K$. Define the cumulative number of edges preceding group $i$ as
\begin{equation}
M_{i-1}
=
\sum_{h=1}^{i-1} m_h,
\qquad
M_0=0.
\end{equation}
Under one-shot decoding, the edges of group $i$ therefore occupy positions
$M_{i-1}+1,\ldots,M_{i-1}+m_i$, giving
\begin{equation}
\label{eq:app_one_error}
\mathcal{E}_{\mathrm{one}}
=
\sum_{i=1}^{K}
\sum_{j=1}^{m_i}
\epsilon_{M_{i-1}+j}.
\end{equation}
Under source-group decoding, each group starts a new autoregressive generation,
so that
\begin{equation}
\label{eq:app_group_error}
\mathcal{E}_{\mathrm{group}}
=
\sum_{i=1}^{K}
\sum_{j=1}^{m_i}
\epsilon_j.
\end{equation}

Since $M_{i-1}+j\geq j$, monotonicity in
Eq.~\ref{eq:app_length_assumption} immediately gives
\begin{equation}
\label{eq:app_grouped_error}
\mathcal{E}_{\mathrm{group}}
\leq
\mathcal{E}_{\mathrm{one}}.
\end{equation}
Hence, in the position-sensitive regime, restarting autoregressive decoding
across source groups weakly reduces the cumulative error component associated
with output position.

\subsection{Bound on Position-Dependent Error Reduction}

We now strengthen the monotonicity assumption by bounding the growth of the
position-dependent error. Let
\begin{equation}
\label{eq:app_increment_bound}
0
\leq
\alpha
\leq
\epsilon_{j+1}-\epsilon_j
\leq
\beta
\end{equation}
for all relevant decoding positions $j$. Here, $\alpha$ and $\beta$ bound the
per-position increase in expected edge-level error.

Define the reduction induced by source grouping as
\begin{equation}
\Delta\mathcal{E}
=
\mathcal{E}_{\mathrm{one}}
-
\mathcal{E}_{\mathrm{group}}.
\end{equation}
Using Eqs.~\ref{eq:app_one_error} and
\ref{eq:app_group_error},
\begin{equation}
\Delta\mathcal{E}
=
\sum_{i=1}^{K}
\sum_{j=1}^{m_i}
\left(
\epsilon_{M_{i-1}+j}-\epsilon_j
\right).
\end{equation}
For every term,
\begin{equation}
\epsilon_{M_{i-1}+j}-\epsilon_j
=
\sum_{r=j}^{M_{i-1}+j-1}
\left(
\epsilon_{r+1}-\epsilon_r
\right).
\end{equation}
Applying Eq.~\ref{eq:app_increment_bound} yields
\begin{equation}
\alpha M_{i-1}
\leq
\epsilon_{M_{i-1}+j}-\epsilon_j
\leq
\beta M_{i-1}.
\end{equation}
Summing over all $m_i$ edges in group $i$ gives
\begin{equation}
\alpha m_i M_{i-1}
\leq
\sum_{j=1}^{m_i}
\left(
\epsilon_{M_{i-1}+j}-\epsilon_j
\right)
\leq
\beta m_i M_{i-1}.
\end{equation}
Summing again over all source groups,
\begin{equation}
\alpha
\sum_{i=1}^{K}
m_i M_{i-1}
\leq
\Delta\mathcal{E}
\leq
\beta
\sum_{i=1}^{K}
m_i M_{i-1}.
\end{equation}
Because
\begin{equation}
\sum_{i=1}^{K} m_i M_{i-1}
=
\sum_{h<i}m_hm_i
=
\frac{1}{2}
\left(
m^2-\sum_{i=1}^{K}m_i^2
\right),
\end{equation}
we obtain
\begin{equation}
\boxed{
\frac{\alpha}{2}
\left(
m^2-\sum_{i=1}^{K}m_i^2
\right)
\leq
\Delta\mathcal{E}
\leq
\frac{\beta}{2}
\left(
m^2-\sum_{i=1}^{K}m_i^2
\right)
}
\label{eq:app_error_reduction_bound}
\end{equation}
with
\[
\Delta\mathcal{E}
=
\mathcal{E}_{\mathrm{one}}
-
\mathcal{E}_{\mathrm{group}}.
\]

Equation~\ref{eq:app_error_reduction_bound} makes explicit how the benefit of
restarting decoding depends on the edge distribution across source groups. If
$K=1$, then $\sum_i m_i^2=m^2$ and the bound correctly gives zero reduction.
For a fixed total number of edges $m$, the quantity
\begin{equation}
m^2-\sum_i m_i^2
\end{equation}
is larger when edges are distributed more evenly across groups, indicating a
larger potential reduction in position-dependent error.

For approximately balanced groups, $m_i\approx m/K$, giving
\begin{equation}
\sum_{i=1}^{K}m_i^2
\approx
\frac{m^2}{K}.
\end{equation}
The lower bound then becomes
\begin{equation}
\label{eq:app_balanced_bound}
\Delta\mathcal{E}
\gtrsim
\frac{\alpha m^2}{2}
\left(
1-\frac{1}{K}
\right).
\end{equation}
Under this stylized regime, the potential benefit of source-group restarting
therefore increases with graph size and with the extent to which the full edge
set is distributed across multiple decoding groups.

\paragraph{Scope of the analysis.}
The analysis separates model-independent structural guarantees from
model-dependent conclusions. The first two results are deterministic under the
serialization setting defined above. Source grouping preserves the complete
target topology while removing cross-group serialization ambiguity, and it
strictly shortens the maximum decoding horizon whenever multiple nonempty
source groups are present. These properties do not depend on any assumption
about VLM error behavior.

The subsequent error analysis is conditional on the position-sensitive
decoding model. In particular, the inequality
$\mathcal{E}_{\mathrm{group}}\leq\mathcal{E}_{\mathrm{one}}$ requires the
assumption that expected edge-level error is nondecreasing with decoding
position, while the quantitative error-reduction bound further assumes bounded
per-position error growth. Prior work provides empirical motivation for this
regime by showing increased hallucination and semantic drift in longer LVLM
generations~\citep{zheng2025lvlms,jiang2026kvsmooth}, but these observations do
not imply that every VLM follows such an error profile. Our derivation therefore
provides unconditional guarantees about representation and decoding length,
together with conditional guarantees about error reduction when longer
autoregressive generations become progressively harder to maintain
consistently.

\section{Additional Experimental Results}

\subsection{Relation-type prediction by domain}
\label{app:type_results}

Table~\ref{tab:type_f1} reports domain-level Typed-Edge F1 on the three
Knossos domains with explicit relation semantics. Across all three typed domains, both Ariadne variants consistently outperform
their corresponding direct backbones. Ariadne-Qwen achieves the highest
Typed-Edge F1 in each domain, with Ariadne-GLM close behind. Circuit remains
the most challenging domain for most methods, while Natural Process generally
shows the strongest performance. The large variation of several closed-source
models across domains further suggests that relation-type prediction remains
sensitive to domain-specific semantics.

\begin{table}[t]
\caption{Typed-Edge F1 (\%) on the three Knossos domains with explicit
relation semantics.}
\label{tab:type_f1}
\centering
\scriptsize
\setlength{\tabcolsep}{3pt}
\renewcommand{\arraystretch}{0.95}

\begin{tabular}{lccc}
\toprule
Method & Natural Process & Circuit & Map Route \\
\midrule

\multicolumn{4}{l}{\textit{Open-source VLMs}} \\
\cmidrule(lr){1-4}
Qwen3-VL-32B      & 49.5 & 30.7 & 69.3 \\
Qwen3-VL-8B       & 41.8 & 24.0 & 55.8 \\
GLM-4.6V-Flash    & 34.9 & 19.8 & 48.3 \\
MiMo-VL-7B-RL     & 34.5 & 21.3 & 57.4 \\
Keye-VL-1.5-8B    & 31.6 & 11.5 & 28.2 \\
InternVL3.5-8B    & 38.9 & 21.2 & 44.1 \\
Ovis2.5-9B        & 24.9 & 17.1 & 43.1 \\
\addlinespace[1pt]

\multicolumn{4}{l}{\textit{Closed-source VLMs}} \\
\cmidrule(lr){1-4}
GPT-4o            & 38.5 & 26.6 & 60.1 \\
GPT-5.6            & 88.9 & 76.6 & 86.5 \\
Opus 5             & 96.3 & 56.1 & 77.1 \\
\addlinespace[1pt]

\multicolumn{4}{l}{\textit{Visual reasoning methods}} \\
\cmidrule(lr){1-4}
Pixel Reasoner      & 16.3 & 11.7 & 21.2 \\
Chain of Region     & 29.8 & 22.9 & 51.2 \\
Speculative Verdict & 43.1 & 20.3 & 58.2 \\
TopoAgent           & 36.3 & 29.6 & 53.4 \\
\addlinespace[1pt]

\multicolumn{4}{l}{\textit{Ours}} \\
\cmidrule(lr){1-4}
\textbf{Ariadne (GLM-based)}
& 95.9 & 82.9 & 87.8 \\
\textbf{Ariadne (Qwen3-based)}
& \textbf{97.3} & \textbf{85.0} & \textbf{89.0} \\
\bottomrule
\end{tabular}
\end{table}




\subsection{Bounding-box and connector localization}
\label{app:geometry_comparison}

We evaluate geometric grounding on a fixed 120-image subset of the Knossos
test set, with 20 images per domain and difficulty proportions approximately
preserved. The subset contains 1,548 reference nodes and 2,759 connector
instances corresponding to typed edges. Bounding-box evaluation uses each
model's predicted nodes, whereas connector localization is conditioned on
reference nodes and edges to isolate geometric localization quality from
topology errors.

For bounding boxes, predictions are matched to reference nodes by normalized
visible names using one-to-one maximum-IoU assignment. We report mean IoU
(mIoU) over matched nodes and recall at IoU $\geq 0.5$ (R@0.5). Connector
localization is evaluated using the mean L2 distance between the five predicted
and reference points along each connector. All distances are measured in the
original image coordinate system.

\begin{table}[t]
\centering
\small
\caption{Geometric grounding on the fixed 120-image Knossos subset.
Box mIoU and R@0.5 are percentages; connector L2 is measured in
original-image pixels.}
\label{tab:geometry_matched120}
\setlength{\tabcolsep}{6pt}
\begin{tabular}{lrrr}
\toprule
Method
& Box mIoU $\uparrow$
& R@0.5 $\uparrow$
& Connector L2 $\downarrow$ \\
\midrule
GPT-5.6
& 81.81 & 93.73 & 11.34 \\
Ariadne-GLM
& 86.42 & 96.96 & 6.14 \\
Ariadne-Qwen3
& \textbf{96.59} & \textbf{97.55} & \textbf{4.13} \\
\bottomrule
\end{tabular}
\end{table}

Table~\ref{tab:geometry_matched120} shows that both Ariadne variants improve
geometric grounding over GPT-5.6, with Ariadne-Qwen3 performing best overall.
Its box mIoU increases from 81.81\% to 96.59\%, while connector localization
error decreases from 11.34 to 4.13 pixels. Ariadne-GLM also substantially
reduces connector localization error while improving bounding-box localization.

\begin{table}[t]
\centering
\small
\caption{Per-domain geometric grounding on the same 120-image subset.
Box mIoU is reported in percent and connector L2 in pixels.}
\label{tab:geometry_matched120_domains}
\setlength{\tabcolsep}{3.5pt}
\resizebox{\linewidth}{!}{%
\begin{tabular}{lrrrrrr}
\toprule
& \multicolumn{3}{c}{Box mIoU $\uparrow$}
& \multicolumn{3}{c}{Connector L2 $\downarrow$} \\
\cmidrule(lr){2-4}\cmidrule(lr){5-7}
Domain
& GPT-5.6 & Ariadne-GLM & Ariadne-Qwen3
& GPT-5.6 & Ariadne-GLM & Ariadne-Qwen3 \\
\midrule
Food Web
& 88.71 & 86.20 & 96.06
& 17.54 & 6.80 & 5.74 \\
Network
& 91.78 & 92.91 & 98.55
& 7.51 & 4.07 & 4.14 \\
Workflow
& 95.15 & 84.62 & 97.75
& 6.80 & 4.78 & 2.86 \\
Natural Process
& 67.74 & 85.78 & 94.94
& 6.68 & 6.09 & 3.37 \\
Circuit
& 69.12 & 82.98 & 93.49
& 20.24 & 8.36 & 4.67 \\
Map Route
& 69.37 & 85.29 & 99.19
& 8.27 & 8.11 & 4.00 \\
\bottomrule
\end{tabular}
}
\end{table}

The domain-level results in
Table~\ref{tab:geometry_matched120_domains} show that the gains are broadly
consistent across diagram types. Ariadne-Qwen3 achieves the highest box mIoU
in all six domains and lower connector localization error than GPT-5.6
throughout. The largest reductions in connector error occur on Food Web and
Circuit, which are also among the most challenging domains for connector
localization.

\begin{figure}[t]
\centering
\includegraphics[width=\linewidth]{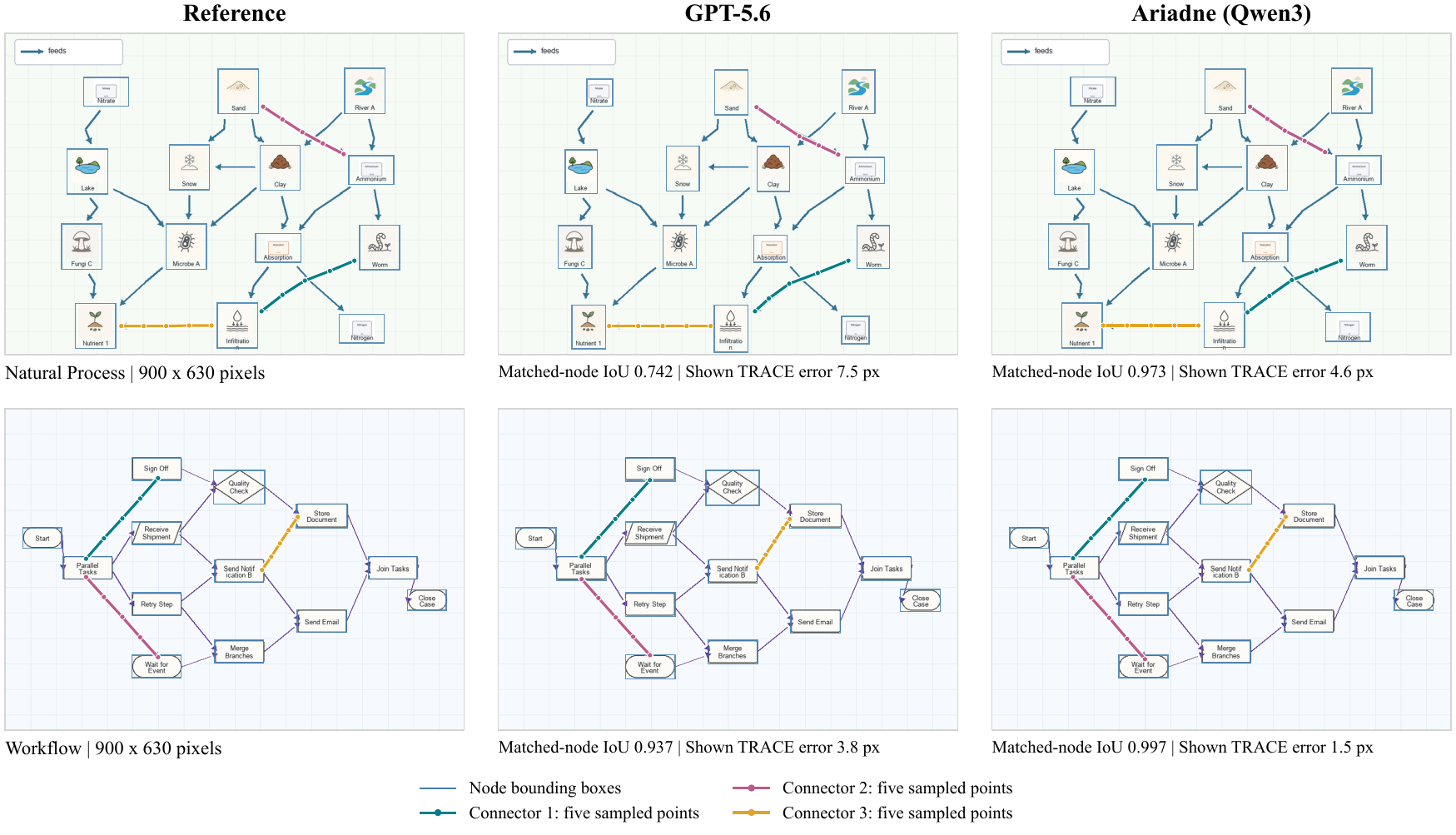}
\caption{Geometric grounding examples on two original Knossos test images.
Boxes show node localization, while colored polylines and markers show
reference-conditioned connector predictions. No predicted coordinates are
manually adjusted.}
\label{fig:geometry_overlay}
\end{figure}

Figure~\ref{fig:geometry_overlay} provides qualitative examples of the same
comparison.

\subsection{Node extraction by domain}
\label{app:node_results}

Table~\ref{tab:node_by_domain} reports the complete domain-level Node F1
results underlying the macro-averages summarized in the main paper, together
with results on the external TopoBench-180 test set. The Knossos average gives
equal weight to all six domains. Across Knossos, node extraction is consistently strong for most models, with
Food Web, Network, Workflow, and Natural Process generally approaching
ceiling performance. Circuit is noticeably more challenging, producing the
largest drop in Node F1 for many baselines, while Map Route also shows somewhat
greater variation. Both Ariadne variants remain stable across domains and
achieve the highest Knossos average of 97.8\%. On TopoBench-180, however,
general-purpose VLMs often retain stronger node-level generalization than the
task-adapted Ariadne node predictors. This domain-level breakdown further
supports the observation that node-specific adaptation improves in-domain
accuracy but may introduce some specialization to the synthetic training
distribution.

\begin{table*}[t]
\caption{Node F1 (\%) across the frozen \textsc{Knossos} internal test set and
TopoBench-180. Avg. is the unweighted macro-average of the six \textsc{Knossos}
domain-level scores. Dashes indicate results that are not available.}
\label{tab:node_by_domain}
\centering
\small
\resizebox{\textwidth}{!}{%
\begin{tabular}{lccccccccc}
\toprule
& \multicolumn{7}{c}{\textsc{Knossos}}
& \multicolumn{2}{c}{TopoBench-180} \\
\cmidrule(lr){2-8}
\cmidrule(lr){9-10}
Method
& Food Web
& Network
& Workflow
& Natural Process
& Circuit
& Map Route
& Avg.
& Web-style
& Network-style \\
\midrule
\multicolumn{10}{l}{\textit{Open-source VLMs}} \\
\cmidrule(lr){1-10}
Qwen3-VL-32B
& 99.3 & 99.5 & 99.4 & 97.9 & 87.8 & 93.9 & 96.3 & 96.3 & 87.5 \\
Qwen3-VL-8B
& 99.3 & 99.6 & 99.4 & 96.0 & 86.3 & 90.9 & 95.2 & 95.4  & 88.9 \\
GLM-4.6V-Flash
& 98.9 & 98.7 & 99.0 & 98.0 & 85.4 & 90.8 & 95.1 & 94.3 & 86.3 \\
MiMo-VL-7B-RL
& 87.0 & 98.6 & 92.3 & 88.7 & 83.3 & 90.4 & 90.0 & 73.7 & 81.2 \\
Keye-VL-1.5-8B
& 97.0 & 98.4 & 96.4 & 92.5 & 80.8 & 87.5 & 92.1 & 86.7 & 83.9 \\
InternVL3.5-8B
& 99.2 & 99.0 & 98.5 & 96.7 & 80.6 & 89.2 & 93.9 & 94.7 & 86.4 \\
Ovis2.5-9B
& 96.7 & 99.1 & 98.6 & 97.3 & 80.2 & 90.9 & 93.8 & 80.1 & 78.6 \\
\addlinespace[3pt]
\multicolumn{10}{l}{\textit{Closed-source VLMs}} \\
\cmidrule(lr){1-10}
GPT-4o
& 99.9 & 99.1 & 99.4 & 98.9 & 89.2 & 93.0 & 96.6 & 93.3 & 84.1 \\
GPT-5.6
& 99.9 & 99.2 & 99.2 & 98.8 & 88.9 & 94.7 & 96.8 & \textbf{96.4} & 87.5 \\
Opus 5
& 99.9 & 99.9 & 99.9 & 99.2 & 90.5 & 95.1 & 97.4 & 93.6 & \textbf{88.1} \\
\addlinespace[3pt]
\multicolumn{10}{l}{\textit{Visual reasoning methods}} \\
\cmidrule(lr){1-10}
Pixel Reasoner
& 99.8 & 99.8 & 99.2 & 91.0 & 82.3 & 90.0 & 93.7 & 75.2 & 71.4 \\
Chain of Region
& 99.8 & 98.6 & 99.5 & 99.2 & 87.5 & 89.2 & 95.6 & 95.6 & 86.8 \\
Speculative Verdict
& 99.6 & 99.8 & 99.3 & 99.2 & 85.5 & 91.0 & 95.7 & 93.6 & 83.2 \\
TopoAgent
& 99.9 & 99.6 & 99.0 & 98.4 & 88.8 & 92.4 & 96.4 & 95.6 & 87.0 \\
\addlinespace[3pt]
\multicolumn{10}{l}{\textit{Ours}} \\
\cmidrule(lr){1-10}
\textbf{Ariadne (GLM-based)}
& 99.7 & 99.5 & 99.8 & 99.2 & 93.4 & 95.0 & \textbf{97.8} & 94.6 & 82.9 \\
\textbf{Ariadne (Qwen3-based)}
& 99.7 & 99.7 & 99.9 & 99.2 & 93.6 & 94.9 & \textbf{97.8} & 92.2 & 79.3 \\
\bottomrule
\end{tabular}
}
\end{table*}

\subsection{Protocol efficiency ablation}
\label{app:protocol_efficiency}

We compare the inference cost of one-step generation and staged decomposition
on the same 30 Knossos test images, with 10 images from each difficulty level.
For each backbone, both protocols use the same model configuration. One-step
generation requires a single model invocation per image, whereas staged
inference performs one node-prediction call followed by one call for each
active source group. Table~\ref{tab:protocol_efficiency} reports mean per-image inference cost.
GPT-5.6 latency, token usage, and API cost are obtained directly from archived
request metadata. For GLM-4.6V-Flash, latency is reconstructed from consecutive
completion timestamps within each continuously running GPU shard, excluding
model loading. The comparison is therefore paired within each backbone;
absolute latency between local GLM inference and the GPT-5.6 API should not be
interpreted as a hardware-versus-service benchmark.

\begin{table*}[t]
\caption{Inference efficiency on the same 30-image test slice
(10 simple, 10 medium, and 10 difficult). S/M/D denote mean latency
(seconds per image) by difficulty; Overall is the mean across all 30 images.}
\label{tab:protocol_efficiency}
\centering
\small
\resizebox{\textwidth}{!}{%
\begin{tabular}{llrrrrrrrr}
\toprule
Backbone & Protocol & Calls/img & API calls/img
& S & M & D & Overall & Tokens/img & Cost/img (\$) \\
\midrule
GLM-4.6V-Flash & One-step
& 1.00 & 0.00 & 61.81 & 106.00 & 153.92 & 107.24 & -- & 0.000 \\
GLM-4.6V-Flash & Staged
& 5.23 & 0.00 & 83.39 & 126.86 & 204.33 & 138.19 & -- & 0.000 \\
\addlinespace
GPT-5.6 & One-step
& 1.00 & 1.00 & 13.89 & 26.98 & 28.90 & 23.26 & 3,843 & 0.052 \\
GPT-5.6 & Staged
& 5.23 & 5.23 & 21.99 & 34.55 & 54.81 & 37.12 & 10,747 & 0.147 \\
\bottomrule
\end{tabular}
}
\end{table*}

Staged inference uses $5.23\times$ as many model invocations on this slice.
Relative to one-step generation, mean latency increases by 28.9\% for
GLM-4.6V-Flash and 59.6\% for GPT-5.6. For GPT-5.6, token usage increases by
$2.80\times$, with a comparable increase in estimated API cost. The overhead
also grows with diagram difficulty because larger node inventories induce more
source-group calls. These results quantify the main efficiency trade-off of
staged decomposition alongside its accuracy gains reported in the main
ablation.

\newpage
\section{Additional Reproducibility Material}

\subsection{Training and inference configuration}
\label{app:training_configuration}

Both Ariadne instantiations use the same data, optimization protocol, prompts,
and output schemas. Table~\ref{tab:training_configuration} summarizes the
stage-specific training hyperparameters and adapter initialization.
Edge and trace predictors each initialize from the completed
node adapter but are trained independently.

\begin{table}[t]
\centering
\small
\setlength{\tabcolsep}{4pt}
\caption{Stage-specific supervised-training configuration. ``Initialization''
denotes the adapter state from which each specialist is trained. Steps are
optimizer steps after gradient accumulation.}
\label{tab:training_configuration}
\begin{tabular}{lrrrl}
\toprule
Stage & Epochs & Steps & Learning rate & Initialization \\
\midrule
NODE   & 1 & 1,125 & $5\!\times\!10^{-5}$ & Pretrained VLM \\
EDGE   & 1 & 5,960 & $1\!\times\!10^{-5}$ & Node predictor adapter \\
TRACE  & 1 & 5,801 & $1\!\times\!10^{-5}$ & Node predictor adapter \\
\bottomrule
\end{tabular}
\end{table}

All trainable Ariadne predictors and same-backbone direct full-graph baselines
are trained with QLoRA using 4-bit NF4 quantization, double quantization, and
bfloat16 computation. We use LoRA rank $r=16$, scaling factor $\alpha=32$, and
dropout $0.05$. Adapters are applied to the language attention and MLP
projections, the corresponding vision projections, and the vision--language
merger. Training optimizes assistant-response tokens only, while image, system,
and user-prompt tokens are masked from the supervised cross-entropy loss. We
use paged 8-bit AdamW with zero weight decay, gradient clipping at 1.0, cosine
learning-rate decay, 3\% linear warmup, and random seed 7. Gradient
checkpointing is enabled. The effective batch size is 16 with a per-device batch size of one. All Ariadne
predictors and direct full-graph baselines are trained on four NVIDIA A100 GPUs
with four gradient-accumulation steps. We train direct full-graph baselines
using GLM-4.6V-Flash (9B) and Qwen3-VL-8B with the same corresponding backbones,
training data, and backbone-specific optimization settings as their Ariadne
counterparts. Model evaluation is performed after training on the frozen
evaluation splits rather than during the training loop.

Inference uses deterministic greedy decoding with one beam. For Ariadne, the
node and edge predictors use maximum generation budgets of 2,048 tokens, while
the trace predictor uses 1,024 tokens. Generation terminates early upon
producing the corresponding closing tag
(\texttt{</NODES>}, \texttt{</EDGE\_GRAPH>}, or \texttt{</TRACES>}).
The predicted node inventory is partitioned into source groups of target size
three, while every group retains the complete predicted inventory as its
candidate target set. The direct full-graph baselines instead generate the
complete node inventory and edge structure within a single response.
Evaluation loads model weights in 4-bit mode, and identical frozen test
examples are used for the GLM- and Qwen3-based Ariadne variants and their
corresponding direct full-graph baselines.

\subsection{Prompts for topology extraction}
\label{app:training_records}

Ariadne separates topology extraction into a node-inventory predictor followed
by a source-group edge predictor. The two predictors are trained with separate
objectives and independent QLoRA adapters. In particular, edge prediction does
not consume or produce connector traces; the symbolic topology is complete once
the source-group edge predictions are merged. During edge training, the
complete reference node inventory is provided as input, whereas end-to-end
inference uses the inventory predicted by Stage~1. Text enclosed in square brackets below denotes per-example content inserted by the data or evaluation pipeline and is not literal prompt text.

\paragraph{Node inventory prompt.}
The node predictor is restricted to node discovery and geometric localization.
Its output contains visible node labels, diagram-local identifiers, and
bounding boxes, but no relations or connector predictions.

\begin{ariadneoutput}{NODE system prompt}
You are the Node Inventory Predictor. Given a topology diagram
image, detect only the visible nodes and return their ids, names, and
bounding boxes. Do not extract edges, traces, legends, or final
graphs.
\end{ariadneoutput}

\begin{ariadneoutput}{NODE user prompt}
Extract only the visible node inventory from this diagram.

Return only a <NODES> block. Do not output edges or traces.

Rules:
- List every visible node exactly once.
- Use node ids n1, n2, ... in reading order.
- Preserve the visible label text as the node name.
- Each box covers the full visible node region, including its
  symbol and text label.
- Use integer coordinates in [x1,y1,x2,y2] format.
\end{ariadneoutput}

\begin{ariadneoutput}{NODE response schema}
<NODES>
  <NODE id="n1" name="Grass" box="[120,40,200,120]"/>
  <NODE id="n2" name="Rabbit" box="[320,220,410,350]"/>
</NODES>
\end{ariadneoutput}

\paragraph{Source-group edge prompt.}
The edge predictor receives the complete node inventory as target context but
may emit outgoing relations only for the active source group. Because the
source groups partition the predicted node inventory, the resulting edge
multisets can be merged deterministically after independent decoding.

\begin{ariadneoutput}{EDGE system prompt}
You are the Edge Predictor. Extract typed topology relations
for the selected source nodes and output EDGE elements only.
\end{ariadneoutput}

\begin{ariadneoutput}{EDGE user prompt template}
Extract typed topology relations for the selected source nodes.

You are given:
1. the diagram image;
2. the complete fixed node list with id, name, and box; and
3. the current source node group.

Use only the provided node ids and names. Output only relations
whose source is in the current source group. For every positive
relation, output EDGE directly; do not output TRACE. Preserve the
visible target and exact relation type. If one source-target pair
has multiple types, output one EDGE per type and do not merge them.

If an active source has no outgoing relation, output an empty
CHECK_NODE. Use sparse NO_EDGE only for hard negatives such as a
near miss, a crossing without connection, or the wrong direction.
Return only <EDGE_GRAPH> and do not explain.

Complete node list:
[NODE_INVENTORY]

Current source node group:
[ACTIVE_SOURCES]

Rules:
- Put each relation under its source CHECK_NODE.
- Emit every active source CHECK_NODE exactly once.
- EDGE carries source, target, their names, and the typed label.
- Never output TRACE or repeat node boxes in the response.
\end{ariadneoutput}

\begin{ariadneoutput}{EDGE response schema}
<EDGE_GRAPH>
<CHECK_NODE id="n4" name="Sensor">
  <EDGE source="n4" source_name="Sensor"
        target="n6" target_name="Controller"
        type="signal"/>
  <EDGE source="n4" source_name="Sensor"
        target="n8" target_name="Ground"
        type="ground"/>
</CHECK_NODE>
</EDGE_GRAPH>
\end{ariadneoutput}

Direction is represented by the ordered source--target pair.
When a diagram legend defines semantic relation types, \texttt{type}
stores the corresponding relation label; otherwise, it is left empty.
Undirected and bidirectional connectors are represented by reciprocal
ordered edges, while multiple typed relations sharing the same
endpoints remain distinct multiset elements.
\texttt{NO\_EDGE} provides sparse hard-negative supervision and is
never inserted into the predicted topology.

\subsection{Independent geometric grounding prompt}
\label{app:trace_prompt}

Connector geometry is predicted only after the symbolic edge set has been
fixed. A third, independently trained QLoRA adapter receives the diagram,
node inventory, and already-decided edges and predicts a five-point trace for
each relation. The trace predictor cannot add, remove, redirect, or retype an
edge. Each input edge is assigned an immutable \texttt{edge\_id}, which links
the predicted geometry back to its corresponding symbolic relation.

For the geometric-grounding evaluation reported in this paper, the trace
predictor is conditioned on reference nodes and reference edges, isolating
connector-localization quality from upstream topology errors.

\begin{ariadneoutput}{TRACE system prompt}
You are the Ariadne Trace Predictor. Given a diagram, fixed nodes, and
already-decided typed edges, localize each edge with exactly five
source-to-target coordinates. Never revise or emit an EDGE.
\end{ariadneoutput}

\begin{ariadneoutput}{TRACE user prompt template}
Localize the already-decided typed edges in the diagram.

The EDGE decisions are fixed. Do not add, remove, rename, retype,
or emit an EDGE. For each input edge, output exactly one TRACE with
the same edge_id. Each TRACE must contain exactly five points
ordered from the source boundary to the target boundary.
Intermediate points must follow the visible connector rather than
a straight-line guess. Return only <TRACES> and stop after
</TRACES>.

Complete fixed node list:
[NODE_INVENTORY]

Already-decided typed edges:
[EDGE_LIST_WITH_EDGE_IDS]
\end{ariadneoutput}

\begin{ariadneoutput}{TRACE response schema}
<TRACES>
  <TRACE edge_id="e1"
    points="[[180,100],[220,145],[260,190],[300,235],[340,280]]"/>
</TRACES>
\end{ariadneoutput}

\subsection{Direct full-graph baseline prompt}
\label{app:baseline_prompt}

For the general-purpose open- and closed-source VLM baselines,
we use direct full-graph generation. Each model receives the
diagram and produces the complete node inventory and edge set
in a single response, using the same node and edge output
conventions as Ariadne.

\begin{ariadneoutput}{Direct full-graph user prompt}
Extract the complete typed topology graph from this diagram in one
response. First output a <NODES> inventory, then an <EDGE_GRAPH>.
Do not explain.

Rules:
- List every visible node exactly once using ids n1, n2, ... in
  from reading order; preserve its visible name and integer box.
- For every positive relation, output an EDGE with the correct
  source, target, and relation type when defined.
- Inspect every connector separately and determine direction only
  from visible arrowheads, never from layout or semantics.
- Preserve the exact domain-specific relation type and all
  multiple types between the same endpoint pair.
- Put each relation under its source CHECK_NODE and cover every
  visible connector.
- Encode a one-arrow connector as one ordered relation.
- Encode a two-arrow or no-arrow connector as two reciprocal
  ordered relations; do not confuse bidirectional and undirected.
- Return only the structured <NODES> and <EDGE_GRAPH> blocks.
\end{ariadneoutput}

\end{document}